\documentclass{article}
\PassOptionsToPackage{dvipsnames,table,xcdraw}{xcolor}

\newcommand{\curve}{\Phi}

\newcommand{\factorsize}[1]{\lvert \mathcal{D}_{#1} \rvert}

\newcommand{\dD}[1]{ \delta \mathcal{D}_{#1} }
\newcommand{\D}[1]{\mathcal{D}_{#1} }
\newcommand{\baseD}{\mathcal{D}}
\newcommand{\SR}[1]{S\big(\pi({#1})\big)}

\newcommand{\ACRO}{{\textbf{FSC}}\xspace}
\newcommand{\target}{\mathcal{E}}

\usepackage[preprint]{corl_2025}

\usepackage{mathrsfs}
\usepackage{amsmath,amssymb,mathtools}
\usepackage{enumitem}
\usepackage{wrapfig}
\usepackage{pifont}
\usepackage{graphicx,scalerel}
\usepackage{booktabs}
\usepackage{multirow}
\usepackage{adjustbox}
\usepackage{xspace}
\usepackage{float}
\usepackage{subcaption}
\usepackage{algorithm,algpseudocode}
\usepackage[capitalise,nameinlink]{cleveref}
\usepackage[font=small,labelfont=bf]{caption}
\usepackage{relsize}
\usepackage{tikz}

\definecolor{princetonorange}{rgb}{0.949,0.522,0.0}
\definecolor{methodblue}{HTML}{D4A2D9}
\definecolor{methodblueborder}{HTML}{936099}

\newcommand\circWithBorder[3][3pt]{%
  \tikz[baseline=-0.6ex]%
       \filldraw[fill=#2, draw=#3, line width=0.4pt] (0,0) circle (#1);\xspace}
\newcommand\bluecircborder{\circWithBorder{methodblue}{methodblueborder}\,}
\newcommand{\tikzdiamond}[3][1.3ex]{%
  \tikz[baseline=0.0ex, x=#1, y=#1, rotate=45]%
    \draw[fill=#2, draw=#3, line width=0.15ex] (0,0) rectangle (1,1);\xspace}
\newcommand\bluefilleddiamond{\tikzdiamond{methodblue}{methodblue}}

\title{Guiding Data Collection via Factored Scaling Curves}

\author{
Lihan Zha$^{1}$\quad Apurva Badithela$^{1}$\quad Michael Zhang$^{1}$\quad \textbf{Justin Lidard}$^{1}$\vspace{0.05in}\\
\textbf{Jeremy Bao}$^{1}$\quad \textbf{Emily Zhou}$^{1}$\quad \textbf{David Snyder}$^{1}$\vspace{0.05in}\\
\textbf{Allen Z. Ren$^{2}$}\quad \textbf{Dhruv Shah}$^{1}$\quad \textbf{Anirudha Majumdar}$^{1}$\vspace{0.05in}\\
 $^1$Princeton University\quad$^2$Physical Intelligence\vspace{0.1in}\\
\href{https://factored-data-scaling.github.io}{\textbf{\textsc{\textcolor{princetonorange}{factored-data-scaling.github.io}}}
}
 }

\begin{document}
\maketitle

\vspace{-25pt}

\begin{abstract}
Generalist imitation learning policies trained on large datasets show great promise for solving diverse manipulation tasks. However, to ensure generalization to different conditions, policies need to be trained with data collected across a large set of environmental factor variations (e.g., camera pose, table height, distractors) --- a prohibitively expensive undertaking, if done exhaustively.
We introduce a principled method for deciding \emph{what} data to collect and \emph{how much} to collect for each factor by constructing \emph{factored scaling curves} (\ACRO{}), which quantify how policy performance varies as data scales along individual or paired factors.
These curves enable targeted data acquisition for the most influential factor combinations within a given budget. 
We evaluate the proposed method through extensive simulated and real-world experiments, across both training-from-scratch and fine-tuning settings, and show that it boosts success rates in real-world tasks in new environments by up to $26\%$ over existing data-collection strategies. 
We further demonstrate how factored scaling curves can effectively guide data collection using an \emph{offline metric}, without requiring real-world evaluation at scale.
\end{abstract}



\begin{figure}[H]
    \centering
    \includegraphics[trim=0 8 0 0, width=1.0\linewidth]{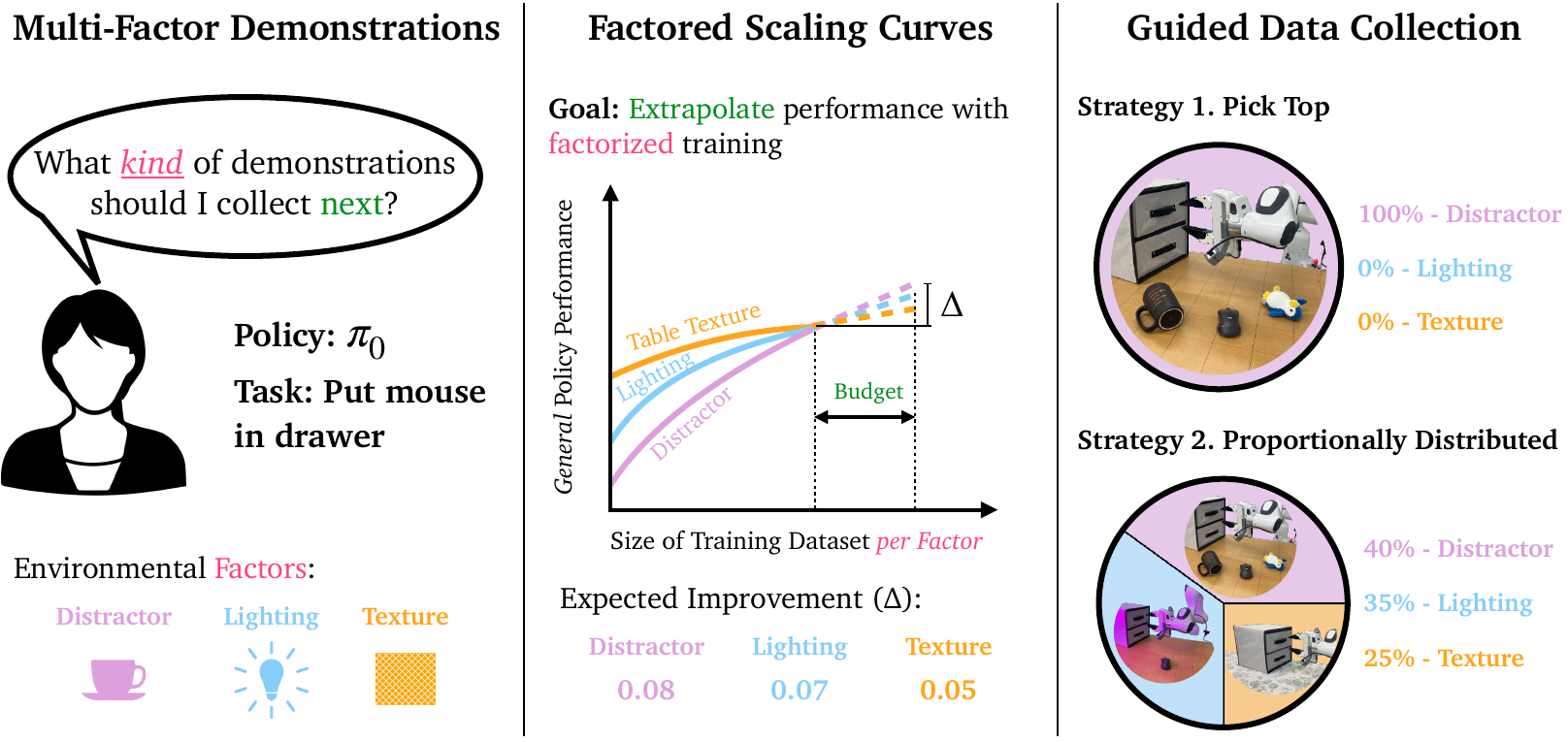}
    \caption{To efficiently collect demonstrations so as to maximize policy performance under a fixed data budget, we propose \emph{factored scaling curves}: a principled tool to quantify how policy performance changes with the quantity of factor data. Based on factored scaling curves, we can allocate the data budget to collecting demonstrations that vary different factors based on their importance.}
    \label{fig:anchor}
\end{figure}

\section{Introduction}
\label{sec:intro}

High-quality teleoperated data has been indispensable for learning many of today's state-of-the-art robot manipulation policies~\cite{brohan2023rt2visionlanguageactionmodelstransfer, black2024pi0visionlanguageactionflowmodel, kim2024openvlaopensourcevisionlanguageactionmodel, geminiroboticsteam2025geminiroboticsbringingai,zhao2024aloha}.
However, robot data collection is prohibitive in time and effort, often requiring more than thousands of hours of human demonstrations~\cite{geminiroboticsteam2025geminiroboticsbringingai,black2024pi0visionlanguageactionflowmodel}. 
Even with large-scale pre-training on existing datasets \cite{khazatsky2024droid, walke2024bridgedatav2datasetrobot, embodimentcollaboration2024openxembodimentroboticlearning, chi2024universalmanipulationinterfaceinthewild}, achieving strong performance in downstream tasks still requires additional in-domain data collection, ranging from a couple of hours to hundreds of hours of effort~\cite{black2024pi0visionlanguageactionflowmodel,kim2025finetuningvisionlanguageactionmodelsoptimizing}. 
For a learned policy to generalize effectively, data collection must also span various environment factor variations, such as differences in table height, object initial state, and camera pose --- this exacerbates the overall effort as collecting data across diverse environment variations requires repeatedly setting up distinct scenarios. Given the substantial data requirements and the high expense of data acquisition, practitioners need an efficient strategy that optimizes policy performance while minimizing human effort and cost.

To this end, we aim to address the question: \emph{given a constrained data budget, which data should be collected to achieve the best policy generalization across varying environmental factor variations (e.g., lighting, backgrounds, camera pose, table height)?} 
A naïve approach might evenly distribute the data budget across all factors, but this is rarely efficient. Not only are there significant hidden costs associated with setting up diverse scenes, but more crucially, the policy’s sensitivity to each factor often varies considerably. 
For instance, if the policy is already robust to camera-pose variations, collecting additional camera-pose demonstrations may provide little incremental benefit, whereas varying table height instead could significantly boost performance.
An effective data collection strategy should prioritize the most impactful factors, and also quantitatively determine the appropriate amount of data to collect for each.

In light of this, we propose a novel framework to systematically prioritize data collection for improving policy generalization across environmental factors. At the core of our approach is the concept of \emph{factored scaling curves}, which model how a policy’s performance improves as additional data is collected involving different factor variations, as shown in~\cref{fig:anchor}. By estimating and \emph{extrapolating} these curves, we can strategically allocate a constrained data budget to the most impactful factors, rather than relying on uniform or heuristic-driven collection.

\paragraph{Statement of Contributions.} We propose a principled robot data collection framework informed by factored scaling curves. Our contributions are as follows: \textbf{(1)} We introduce \emph{factored scaling curves} (\ACRO{}) to quantify how policy performance scales with data for different environmental factors, and show that these curves reliably predict expected policy performance. \textbf{(2)} Building on these curves, we propose a suite of data collection strategies, including top-1 and weighted top-k selection methods that prioritize factors expected to yield the greatest policy performance gains. \textbf{(3)} We validate our framework through extensive experiments in both simulation and real-world robotic manipulation tasks, where we train policies from scratch and fine-tune pre-trained Vision-Language-Action (VLA) models, achieving up to \textbf{26\%} higher success rate than state-of-the-art baselines. \textbf{(4)} We further demonstrate that constructing \ACRO{} solely from policy embedding similarity --- an offline metric that does not require hardware evaluation --- retains almost the same effectiveness in guiding data collection, yielding an extremely lightweight variant of our method. Importantly, each contribution of our framework is general: it applies to \emph{any task} and \emph{any policy backbone}, and can be seamlessly and effectively integrated with existing data collection techniques such as compositional data generation~\cite{gao2024efficientdatacollectionrobotic}.

\vspace{-7pt}
\section{Related Work}
\vspace{-7pt}

\textbf{Theoretical Frameworks for Data Collection.}  Several existing works study dataset construction for improved learning dynamics. For static datasets, coreset selection, optimization, and heuristic tuning~\cite{nguyen2025minibatchcoresetsmemoryefficientlanguage,guruprasad2024benchmarkingvisionlanguage,hejna2024remixoptimizingdatamixtures,hejna2025robotdatacurationmutual,NEURIPS2023_dcba6be9,embodimentcollaboration2024openxembodimentroboticlearning,liu2025regmixdatamixtureregression} find optimal data subsets from larger training sets. However, these approaches assume a fixed, static dataset. By contrast, our objective is to \emph{actively} decide what additional data to gather, akin to active data allocation and learning methods including Bayesian experimental design~\cite{f1168103-f0ea-30c5-aa6c-babe695713d3, 10.1214/ss/1177009939}, information gain maximization~\cite{mackay_1992, houlsby2011bayesianactivelearningclassification}, 
and active learning~\cite{sener2018active}. 
In general, the first two methods require explicit parametric representations of the estimation problem, while the third only chooses the best single arm (i.e., factor). By contrast, our setting seeks to find the best data mixture without overly strong assumptions about the influence mechanism. Additionally, the latter methods often give guarantees via reductions to estimation problems (e.g., \cite{anwar_efficient_2025}) which do not account for the full endogeneity of policy performance with respect to \emph{new} data generation. 

\textbf{Scaling Laws.} Scaling laws quantify model performance improvements with increasing data and compute. Scaling laws have been heavily studied in natural language processing (NLP)~\cite{kaplan2020scalinglawsneurallanguage,openai2024gpt4technicalreport,NEURIPS2020_1457c0d6,hoffmann2022trainingcomputeoptimallargelanguage,grattafiori2024llama3herdmodels} and computer vision~\cite{pmlr-v139-radford21a,Zhai_2022_CVPR,peebles2023scalable,henighan2020scalinglawsautoregressivegenerative}, 
and have seen preliminary investigations in robotics~\cite{sharma2018multiple,kalashnikov2018scalable,cabi2020scalingdatadrivenroboticsreward,walke2024bridgedatav2datasetrobot, zhao2024aloha, embodimentcollaboration2024openxembodimentroboticlearning,lin2024datascalinglawsimitation}. These scaling analyses typically characterize the large-data regime and treat \emph{all} data as a single category. Our approach instead targets the small-data regime and extrapolates scaling curves that quantify the marginal value of adding data for \emph{different factor variations}. This allows for fine-grained analysis to predict which factors will most improve performance.

\textbf{Data Collection Strategies in Robotics.} 
Prior methods offer broad recommendations for collecting higher-quality real-world data~\cite{gao2024efficientdatacollectionrobotic,belkhale2023dataqualityimitationlearning}, but these guidelines remain agnostic to the specific task and policy at hand. A complementary line of research targets efficiency by probing a policy’s failure modes --- through shared-autonomy corrections or compatibility-based selection to gather more informative demonstrations~\cite{liu2022robot,Cui_2023,gandhi2022elicitingcompatibledemonstrationsmultihuman}. Yet, these approaches operate at the \emph{trajectory} level and do not address performance drops stemming from changes in the surrounding environment.
Red-teaming techniques have recently been proposed to estimate a policy’s sensitivity to individual environmental factors and steer data collection accordingly~\cite{majumdar2025predictiveredteamingbreaking}. However, this method does not model how performance will \emph{evolve} as new data are added.
We close these gaps with \emph{factored scaling curves}: a task- and policy-aware framework that predicts performance gains as a function of additional data for each environmental factor. By quantifying the marginal return of collecting more demonstrations along each axis, our method provides principled, budget-aware guidance for prioritizing the most impactful factor variations and thus accelerates real-world policy improvement.
\section{Factored Scaling Curves for Guiding Imitation Data Collection}
\label{sec:method}

Consider the scenario where we have a \emph{pre-trained robot policy} and observe insufficient performance in a target domain. Gathering additional demonstrations for imitation learning can help bridge the gap. We present a data collection strategy that can: (a) determine and prioritize factors for greatest potential improvement, and (b) predict the effect of adding data for a specific factor --- or combination of factors --- on the policy's performance in the target domain.

\subsection{Problem Formulation}
We consider imitation learning policies, either  pre-trained (e.g., on ~\cite{khazatsky2024droid,walke2024bridgedatav2datasetrobot,embodimentcollaboration2024openxembodimentroboticlearning}) or trained from scratch. We assume access to a new set of training demonstrations \(\mathcal{D}\) comprising of variations across $N$ environment factors $\mathcal{F}=\{f_1, f_2, ..., f_N\}$, denoted as 
\begin{equation}
    \mathcal{D} = \mathcal{D}_{\text{nom}} \cup \mathcal{D}_1 \cup \mathcal{D}_2 \cup \dots \cup \mathcal{D}_N,
\end{equation}
where \(\mathcal{D}_{\text{nom}}\) is the set of demonstrations with all environmental factors in a nominal setting (e.g., no distractors, nominal lighting and table texture), and $\mathcal{D}_i$ contains all demonstrations with variations of factor $f_i$ with respect to its nominal value. 
We denote $|\mathcal{D}_i|$ as the number of demonstrations available for factor $f_i$.
A policy trained on dataset $\mathcal{D}$ is denoted as $\pi(\mathcal{D})$, and is evaluated on a target distribution \(\target\) of environments with factor variations unseen in \(\mathcal{D}\). 
The policy’s \emph{overall performance}, denoted $S(\pi(\mathcal{D}))$, is defined as the expected value of a success metric (e.g., partial credit, binary success) on the target distribution \(\target\).
Our goal is to determine how to collect an additional dataset $\Delta \mathcal{D}$, subject to a constraint on the number of additional demonstrations, i.e., $|\Delta \mathcal{D}| \leq K$, where $K$ represents a budget determined by time or data collection cost. The objective is to maximize the performance of the new policy ${\pi}(\mathcal{D} \cup \Delta \mathcal{D})$, trained on the updated dataset: 
\begin{equation}
\label{eq:opt}
    \Delta \mathcal{D} = \arg\max_{\Delta \mathcal{D}} \ S(\pi(\mathcal{D} \cup \Delta \mathcal{D})) \quad \text{s.t.} \quad |\Delta \mathcal{D}|\leq K.
\end{equation}

The additional dataset can be partitioned into subsets corresponding to different factor variations: 
\begin{equation}
    \Delta \mathcal{D} = \Delta \mathcal{D}_1 \cup \Delta \mathcal{D}_2 \cup \dots \cup \Delta \mathcal{D}_N.
\end{equation}
Our focus in solving~\eqref{eq:opt} is to identify \emph{which} factors to prioritize for data collection and \emph{how much} additional data to collect for them, i.e., determining $|\Delta\mathcal{D}_i|$. While this formulation allows for any demonstration collection rule, in this work all demonstrations will vary only one factor at a time.

\subsection{Factored Scaling Curves} 
\label{sec:fsc}
We propose \emph{factored scaling curves} (\ACRO{}) to achieve the aforementioned desiderata. For exposition, we define each curve for an individual factor, and provide extensions to multi-factor settings later in the section. 
For each factor \(f_i\), starting with no corresponding demonstrations, i.e., \(\mathcal{D}\setminus \mathcal{D}_{i}\), we  incrementally add back \(n\) demonstrations \(\dD{i}^{n} \subseteq \D{i}\), and train a policy. Henceforth, denote \(\mathcal{D}^{n}_{i} \coloneqq (\mathcal{D} \setminus \mathcal{D}_{i}) \cup  \dD{i}^n\). 
The \emph{factored scaling curve} \(\curve_{i}: \mathbb{N} \rightarrow \left[0,1\right]\) maps the number of demonstrations of factor \(f_i\) to the policy’s \emph{overall} performance on \(\target\):
\begin{equation}
    \curve_{i}(n) \coloneqq \mathbb{E}_{\D{i}^n \sim \D{i}}\bigg[S\big(\pi(\D{i}^n)\big)\bigg].
\end{equation}

\begin{wrapfigure}[16]{r}{0.47\textwidth}
  \centering
  \vspace{-5pt}
\includegraphics[width=0.47\textwidth]{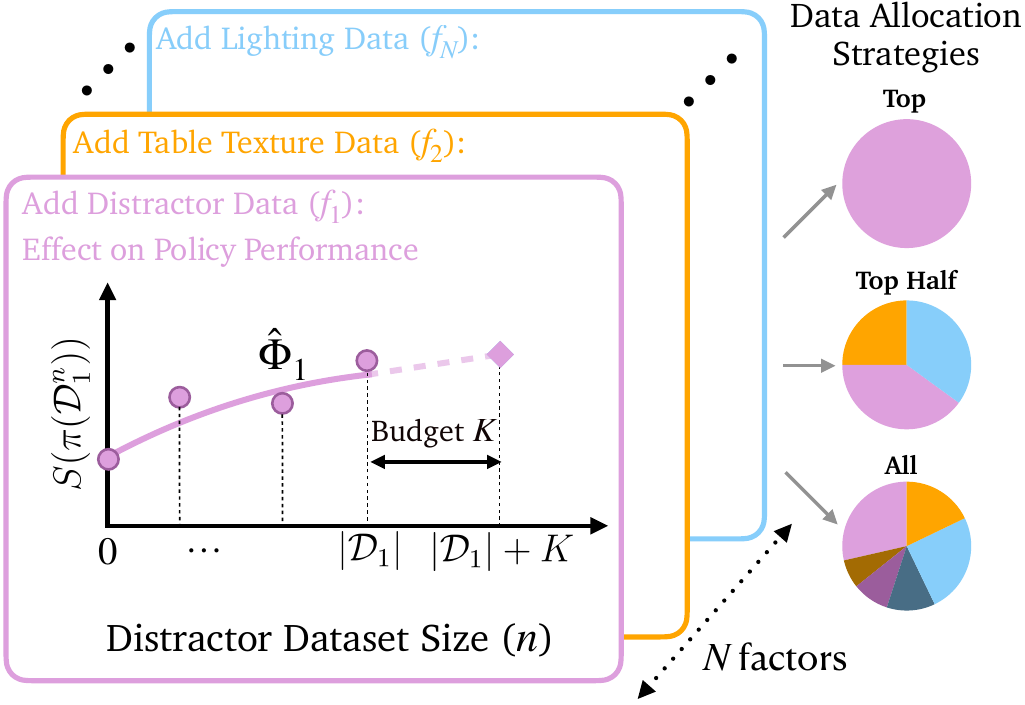}
  \caption{Illustration of factored scaling curves used to inform data allocation. For the distractor factor, \bluecircborder points are used to construct the scaling curve, and \bluefilleddiamond is the predicted policy success rate at \(K\) additional demos of the factor over the initial dataset.}
  \label{fig:method}
  \vspace{-40pt}
\end{wrapfigure}
At \(n = \factorsize{i}\), the scaling curve represents policy performance when using the full available dataset \(\mathcal{D}\) --- comprising of demonstrations from all factors. Note that constructing the curve \emph{does not require gathering additional training demonstrations} beyond the dataset \(\mathcal{D}\). 
Below we summarize some properties of factored scaling curves. First, the discrete derivative quantifies the expected performance gain per additional demonstration, enabling principled ranking of factors. Second, since scaling curves measure a policy's performance in the target domain, they capture how data from one factor affects the policy's \emph{overall} performance including in other factors. Finally, with suitable parametrizations (e.g., fitting a power law), we ensure that the scaling curve captures the saturation effect of adding more data. 

\paragraph{Curve Fitting.} We approximate the factored scaling curve by  training policies \(\pi(\D{i}^k)\) at few equally spaced values of \(k\), and evaluating their performance. This yields points \(\bigl(k,\, \SR{\D{i}^{k}}\bigr)\), which are used to fit a \emph{power-law} model of the factored scaling curve:
\begin{equation}
\hat{\curve}_{i}(n) \coloneqq 1 - a(n+\lvert\baseD \setminus \D{i}\rvert )^b,\qquad a > 0,b < 0,\, \text{ and } n \in \mathbb{N}.
\label{eq:fsc_construction}
\end{equation}
Power laws can effectively model how performance scales with training dataset size in domains such as language modeling~\cite{kaplan2020scaling} and imitation learning~\cite{lin2024datascalinglawsimitation}.
We fit power-law curves in log–log space for numerical stability, following standard practice~\cite{clauset2009power}, and find that as few as four values of \(n\) are often sufficient to obtain a reliable fit empirically.~\cref{fig:method} illustrates the curve construction and its use in predicting the policy's performance if \(K\) additional demonstrations of the factor are gathered.

\paragraph{Proxy Metrics.}
Constructing scaling curves using real-world success rates \(S\) can be expensive in terms of evaluation cost. To address this challenge, we consider other offline metrics \(M\) (e.g.,  embedding similarity~\cite{og_emb_distance,word2vec,majumdar2025predictiveredteamingbreaking}), which do not require evaluating the policies $\pi(\mathcal{D}_i^k)$ on hardware. Generally,  we define factored scaling curves as:
\begin{equation}
    \curve_{i}(n) \coloneqq \mathbb{E}_{\D{i}^n \sim \D{i}}\bigg[M\big(\pi(\D{i}^n)\big)\bigg].
\end{equation}
The resulting performance of the policy trained with additional data is still evaluated according to the gold-standard performance \(S\) in the real-world.  
We show experimental results using embedding-space similarity, denoted \textbf{FSC-Proxy}, in~\cref{sec:emb_distance}.

\paragraph{Factor Combinations.}
Constructing factored scaling curves for each individual factor 
can be expensive in terms of computation and hardware evaluations. Below we discuss how factored scaling curves can be adapted to combine multiple factors into a single scaling curve. We define {Group-t} to be a disjoint partition of \(N\) factors into groups of size \(t\); for example, {Group-2} results in \(\lceil N/2\rceil\) paired combinations. In contrast, {t-wise} refers to all \(\binom{N}{t}\) combinations. 
To balance the expressivity from t-wise and efficiency from Group-t, 
we consider the following options: i) varying individual factors (``{\bf One Factor}"), ii) {2-wise} (``{\bf Pairwise}"): all pairwise factor combinations which results in \(\binom{N}{2}=\tfrac{N(N-1)}{2}\) curves, and iii) Group-2 (``{\bf Group}"): a set of \(\lceil N/2\rceil\) pairwise combinations. The {\bf Pairwise} setting requires more curves than {\bf One Factor} but has greater expressive power, while {\bf Group} requires fewer curves with less expressive power.

\subsection{Data-collection strategy}
With the constructed curves, we now decide which factor(s) to prioritize and how many demos to collect for each factor.
For simplicity, we present the case of \textbf{One Factor} first. The predicted policy performance after adding \(K\) demonstrations of factor \(f_i\) is \(\hat{\curve}_{i}\!\bigl(\lvert\mathcal{D}_{i}\rvert+K\bigr)\). We coarsely approximate the slope of the scaling curve as
\begin{equation}
\label{Eqn:Pij_one}
P^K_{i}\;:=\;\frac{\hat{\curve}_{i}\!\bigl(\lvert\mathcal{D}_{i}\rvert+K\bigr) - \hat{\curve}_{i}\!\bigl(\lvert\mathcal{D}_{i}\rvert\bigr)}{K}.
\end{equation}
Based on \cref{Eqn:Pij_one}, we consider three data collection strategies: {\bf (1) Top:} Identify the top factor with highest \(P^K_{i}\) and allocate the entire budget to it, {\bf (2) Top-Half:} Identify the top half of the factors and allocate budget proportionally, and {\bf (3) All:} Spread the budget over \emph{all} factor combinations in proportion to the respective \(P^K_{i}\). The proportional budget allocation follows:
\( \label{eq:data_alloc_one}
    \lvert\Delta\mathcal{D}_{i}\rvert\;=\; \frac{P^K_{i}}{\sum_{i'} P^K_{i'}}\;K.
\)
Next for \textbf{Pairwise} and \textbf{Group}, similar to the single factor case, we denote the two-factor dataset \(\D{ij} = \D{i} \cup \D{j}\) for factors \(f_i\) and \(f_j\), and define the terms \(\hat{\Phi}_{ij}\) and \(P^K_{ij}\) analogously. The three data collection strategies are defined similarly as with single factor. In the \textbf{Group} setting, the proportional budget allocation strategy for factor combinations is:
\begin{equation}
    \label{eq:data_alloc_group}
    \lvert\Delta\mathcal{D}_{ij}\rvert\;=\; \frac{P^K_{ij}}{\sum_{i',j'} P^K_{i'j'}}\;K,
\end{equation}
with the budget allocated to the individual factors being half of the budget allocated to corresponding factor combination according to \cref{eq:data_alloc_group}. See~\cref{app:data_collect} for details on the \textbf{Pairwise} setting.

\section{Experiments}
\label{sec:result}

We evaluate our proposed method, \ACRO{} (Factored Scaling Curves), alongside \textbf{FSC-Proxy}, which builds FSCs using policy embedding similarity as an offline proxy metric, to address the following questions: (1)~Can our method successfully guide data collection under a fixed data budget to maximize the policy performance? (2)~How well do factored scaling curve extrapolations predict performance with additional data? (3)~How do choices of the prediction strategy and curve construction affect the performance and computation cost?
(4) Can we construct scaling curves using proxy metrics that do not require hardware evaluations while still effectively guiding data collection?

\vspace{-5pt}
\paragraph{Environment Factors.}
We investigate eight factors --- five \emph{visual} (table texture, lighting, camera pose, distractor objects, background) and three \emph{spatial} (table height, object pose, robot initial pose).
Discrete factors (table texture, distractors, background) are drawn from four preset values, whereas continuous factors are sampled uniformly.
See \cref{app:sim_exp_setup} and \cref{app:real_exp_setup} for full distributions and visualizations.
\vspace{-5pt}
\paragraph{Simulation setup.}
We study five simulation tasks in ManiSkill3 \cite{taomaniskill3} on a Franka Panda robot:
\emph{Pick Place}, \emph{Peg Insertion – Visual}, \emph{Peg Insertion – Spatial}, \emph{Pull Cube Tool – Visual}, and \emph{Pull Cube Tool – Spatial}.
Visual tasks vary the five visual factors, and spatial tasks additionally vary the three spatial factors.
All policies are trained with diffusion policy \cite{chi2023diffusionpolicy}. To obtain the factored scaling curve and evaluation results, we evaluate each policy for roughly 4000 trials on different factor values. More details can be found in \cref{app:sim_exp_setup}.
\vspace{-5pt}

\paragraph{Real-world setup.}
We consider two task settings on a Franka Panda robot:
(i) \textbf{fine-tuning VLA}, where we use $\pi_{0}$ as the base model \cite{black2024pi0visionlanguageactionflowmodel} and study three tasks \emph{Fold Towel – Visual}, \emph{Fold Towel – Spatial}, and \emph{Mouse in Drawer};
and (ii) \textbf{train-from-scratch} on \emph{Pick Place} with diffusion policy \cite{chi2023diffusionpolicy}.
We collect training data following the L-shape strategy of \citet{gao2024efficientdatacollectionrobotic}, where each demonstration varies exactly one factor.
For visual experiments, we vary table texture, lighting, camera pose, and distractors. We drop the background variation as we find it has negligible effect in policy performance in our experiment setup.
\emph{Fold Towel – Spatial} and \emph{Mouse in Drawer} additionally vary object and robot poses.
To fit the factored scaling curves and evaluate each policy, we run roughly 15 out-of-distribution trials per policy in which multiple factors are simultaneously varied beyond the training distribution.
Implementation and hardware details are given in \cref{app:real_exp_setup}.

\paragraph{Baselines.} We consider three baseline methods: (1) \textbf{Equal:} Collect an equal number of demonstrations for each factor where we vary exactly one factor value when collecting demos; this is equivalent to the L-shape strategy of \citet{gao2024efficientdatacollectionrobotic}. Outperforming this baseline requires prioritizing the most influential factors. (2)~\textbf{Greedy:} After evaluating the initial policy, we allocate the data budget to the single factor with the lowest success rate. 
(3)~\textbf{Re-Mix:} Following \citet{hejna2024remixoptimizingdatamixtures}, we apply distributionally robust optimization to compute factor weights and construct the initial dataset and collect data in proportion to those weights.

\subsection{How well does \ACRO guide data collection?}

\begin{table*}[h]
\centering
\small
\setlength{\tabcolsep}{5pt}    
\renewcommand{\arraystretch}{1.05}
\caption{\textbf{Evaluating \ACRO in simulation.} We report the average policy success rate trained with additional collected data. \ACRO consistently improves upon the baselines, delivering around $10\%$ improvement on average.}
\begin{tabular}{
  l         
  *{4}{c}   
  @{\hspace{6pt}}  
  *{4}{c}   
}
\toprule
\multicolumn{1}{c}{} &
\multicolumn{4}{c}{$K=20$} &
\multicolumn{4}{c}{$K=100$} \\
\cmidrule(lr){2-5}\cmidrule(lr){6-9}
\textbf{Task}  &
\textbf{\ACRO} & \textbf{Equal} & \textbf{Greedy} & \textbf{Re‑Mix} &
\textbf{\ACRO} & \textbf{Equal} & \textbf{Greedy} & \textbf{Re‑Mix} \\
\midrule
Pick Place &  \textbf{62.0} & 56.1 & 58.7 & 61.6 & 64.4 & 64.3 & \textbf{65.9} & 64.7 \\
Peg Insertion - Visual  &  \textbf{22.2} & 20.2 & 15.5 & 19.2 & \textbf{45.3} & 28.1 & 28.1 & 34.0 \\
Peg Insertion - Spatial &  \textbf{45.5} & 43.8 & 42.0 & 31.7 & \textbf{57.9} & 49.5 & 52.7 & 44.1 \\
Pull Cube Tool - Visual & \textbf{68.4} & 62.7 & 64.3 & 61.9 & \textbf{83.5} & 56.6 & 83.1 & 28.7 \\
Pull Cube Tool - Spatial & \textbf{76.3} & 57.7 & 73.3 & 50.5 & \textbf{83.4} & 78.5 & 62.5 & 64.5 \\
\textbf{Average}  & \textbf{54.9} & 48.1 & 50.8 & 45.0 &
          \textbf{66.9} & 55.4 & 58.5 & 47.2 \\
\bottomrule
\end{tabular}

\label{table:simulation}
\end{table*}
\vspace{8pt}

Simulation results are summarized in \cref{table:simulation}. If not else specified, we adopt the \textbf{Group} construction on the $x$-axis and the \textbf{Top} allocation strategy for the \textbf{FSC} result, which \cref{sec:ablation} later identifies as the best balance between performance and data-collection cost. Results are reported under two data budgets: a small budget ($K=20$) and a large budget ($K=100$).
\ACRO outperforms all baselines in every task except one cell (\emph{Pick Place}, $K=100$), where it is a close second to \textbf{Greedy}, which is otherwise the best-performing baseline on average.
In the challenging, long-horizon task \emph{Pull Cube Tool -- Visual},  \ACRO delivers around $10\%$ improvement over all baselines at $K{=}100$, confirming that the factored scaling curves extrapolate well beyond their fit range and guide data collection effectively. We show visualizations of factored scaling curves in \cref{sec:exp_curve} and \cref{app:additional_curve}. 
Notably, performances of \textbf{Equal} and \textbf{Greedy} are \emph{highly inconsistent} across tasks, and \textbf{Re-Mix} remains consistently weak, whereas \ACRO{} provides stable gains throughout. For example, in the \emph{Pull Cube Tool -- Spatial} task, \textbf{Equal} performs poorly when $K=20$ but has reasonable performance when $K=100$. However, in the \emph{Peg Insertion -- Visual} tasks, this trend is reversed. The same observation holds for another heuristic baseline \textbf{Greedy}, where it consistently has unsatisfactory performance in the \emph{Peg Insertion - Visual} task and inconsistent performance in \emph{Pull Cube Tool - Spatial}. 
\begin{figure}
    \centering
    \includegraphics[width=1.0\linewidth]{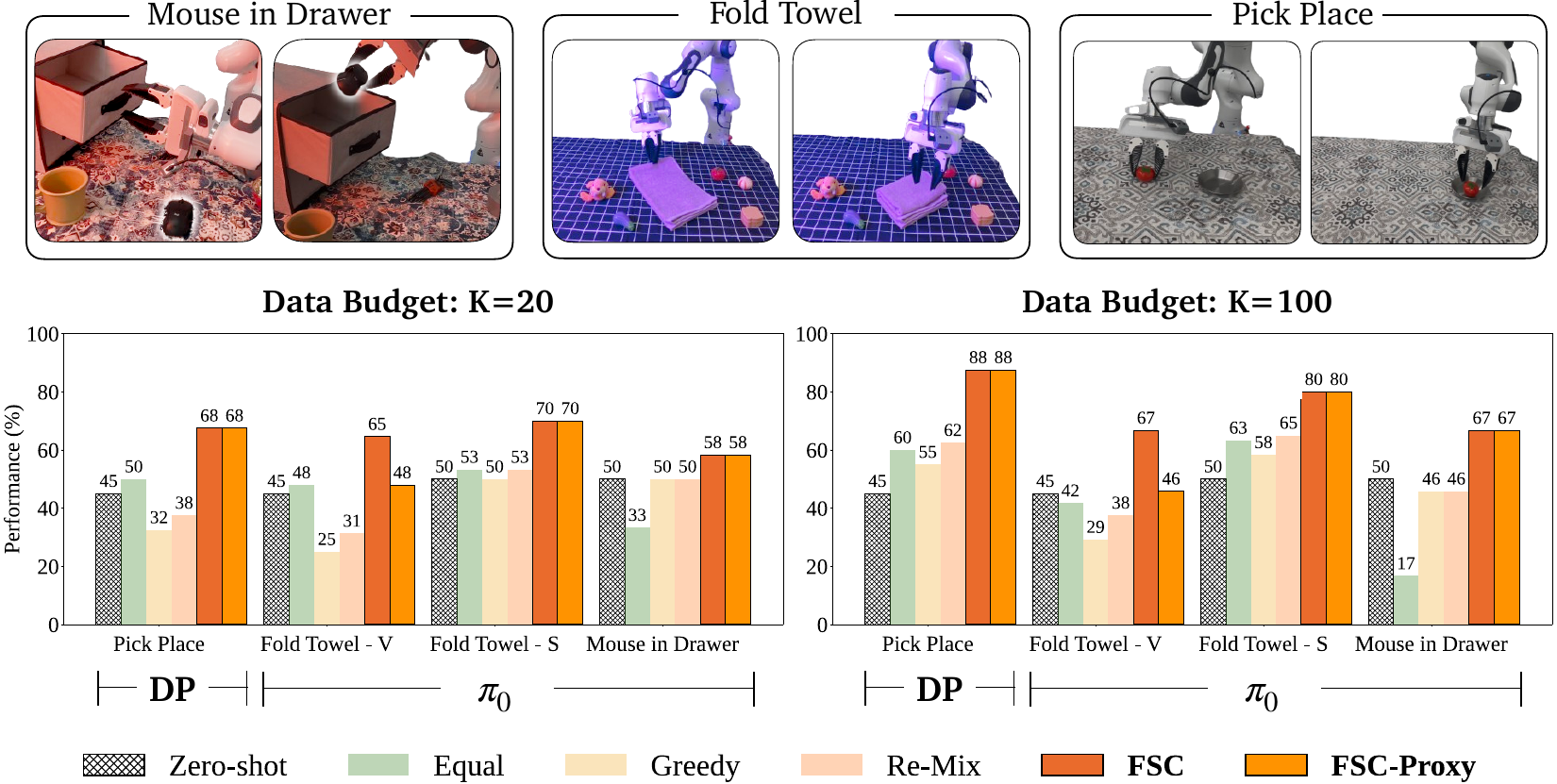}
    \caption{\textbf{Evaluating \ACRO in the real world.} We visualize the task rollouts and report the average policy success rate trained with additional collected data. For \emph{pick-place} task, we train the policies with diffusion policy. For all other experiments, we obtain policies by fine-tuning $\pi_0$. \ACRO achieves the best performance in all tasks, achieving up to \textbf{26\%} more improvement over all baseline methods. Compared to the zero-shot setting, fine-tuning $\pi_0$ with \ACRO{} yields up to $30\%$ success rate improvement. \textbf{FSC-Proxy} achieves nearly the same high success rate as \ACRO{} while eliminating the need for any on-hardware policy execution.}
    \vspace{-10pt}
    \label{fig:real_exp}
\end{figure}

\paragraph{Real-world experiments.} \cref{fig:real_exp} shows that real-world results closely match findings in simulation: \ACRO{} outperforms every baseline by a wide margin. In the \textbf{fine-tuning VLA} setting, \ACRO{} raises success on demanding long-horizon tasks ---\emph{Fold Towel} and \emph{Mouse in Drawer} --- by up to \textbf{25\%} and \textbf{21\%} respectively over the strongest baseline. Increasing the budget from $K=20$ to $K=100$ brings great gains for \ACRO{}, whereas \textbf{Equal} and \textbf{Greedy} improve only marginally or even degrade. A similar pattern emerges in the \emph{Pick Place} task trained with diffusion policy, where \ACRO{} achieves up to a \textbf{26\%} advantage. These results confirm that \ACRO{} not only guides data collection effectively but also generalizes across real-world settings of varying task difficulty and policy type.

\subsection{How well do factored scaling curves predict performance with additional data?}
\label{sec:exp_curve}

\begin{figure}
    \centering
    \includegraphics[width=1.0\linewidth]{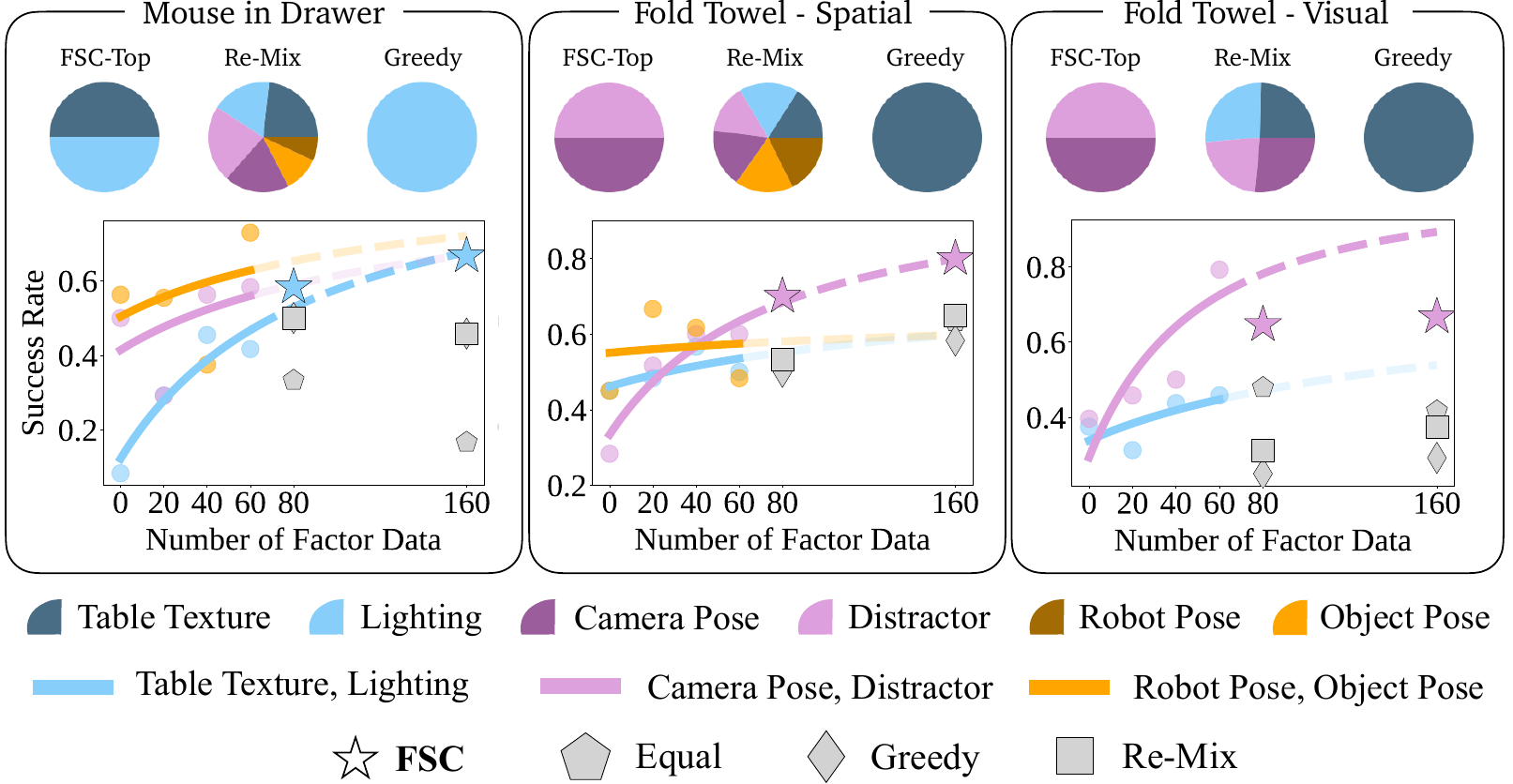}
    \caption{\textbf{Visualizing factored scaling curves for real world fine-tuning $\pi_0$ experiments}. Solid lines are factored scaling curves we construct based on the initial dataset, and dashed lines are the extrapolations that predicts how policy performance change with additional factor data. Based on the \textbf{Top} strategy, \ACRO{} suggests picking the curve with the highest slope, shown in blue (left), purple (middle) and purple (right). Factored scaling curves can \emph{accurately predict how policy performance changes with additional factor data}, thus able to provide informed data collection strategies. We also visualize how different methods allocate data collection budget to the factors in the top pie charts.}
\vspace{-15pt}
\label{fig:real_finetune_curve}
\end{figure}

We visualize our factored scaling curve for the real-world fine-tuning tasks in~\cref{fig:real_finetune_curve}. As we adopt the \textbf{Top} strategy for data collection, we are essentially collecting data for the factor with the highest expected improvement. 
In \emph{Mouse in Drawer}, the (\emph{Table Texture, Lighting}) curve offers the highest expected improvement, so we allocate the entire additional data budget to that factor pair (blue stars at $n{=}80$ and $n{=}160$, matching the $K{=}20$ and $K{=}100$ settings). Even though the curve is fitted only on $n{=}0\text{–}60$, its extrapolation matches the actual performance almost perfectly. The same holds for \emph{Fold Towel – Spatial}: adding data for (\emph{Camera Pose, Distractor}) improves success rate exactly as predicted. This accuracy underpins \ACRO{}’s large margins over baselines. Furthermore, \ACRO is robust to real evaluation noise. In \emph{Fold Towel – Visual} an outlier at $n{=}60$ slightly distorts the fit, yet \ACRO{} still selects the right factor; factor \emph{combinations} are helpful here since they widen the data range and improve the signal-to-noise ratio.

In~\cref{fig:real_finetune_curve}, the pie plots beside each curve show weights allocated to each factor. \ACRO{} allocates the entire budget to the best factor group and is then split evenly inside that group (e.g., $50\%$ each to table texture and lighting). \textbf{Greedy} often misallocates budget to insignificant factors. \textbf{Re-Mix} consistently performs poorly because it either learn near-uniform weights or concentrate on irrelevant factors --- it produces near-uniform weights for the \emph{Fold Towel - Spatial} and \emph{Fold Towel - Visual} task, while not prioritizing the important factors enough (i.e., lighting and table texture) in the \emph{Mouse in Drawer} task.

Interestingly, the pre-trained $\pi_{0}$ is still vulnerable to visual perturbations. Across all three tasks, additional demonstrations that vary \emph{visual} factors deliver the greatest improvements in success rate. In contrast, spatial robustness depends more on the \emph{diversity} than the \emph{quantity} of spatial data: enlarging the set of robot- or object-pose variations produces little further gain, indicating that the initial dataset already captures spatial variation well. This pattern matches the findings of ~\citet{xue2025demogen}.

\subsection{What is the best curve construction choice and prediction strategy?}
\label{sec:ablation}

\begin{table*}[h]
\centering
\small
\setlength{\tabcolsep}{5pt}    
\renewcommand{\arraystretch}{1.05}
\caption{Comparisons of different curve construction choices. The \textbf{Group} setting achieves high performance with lowest computational costs.}

\begin{tabular}{
  l         
  *{4}{c}   
  @{\hspace{6pt}}  
  *{4}{c}   
}
\toprule
\multicolumn{1}{c}{} &
\multicolumn{4}{c}{$K=20$} &
\multicolumn{4}{c}{$K=100$} \\
\cmidrule(lr){2-5}\cmidrule(lr){6-9}
\textbf{Task}&
\textbf{One Factor} & \textbf{Pairwise} & \textbf{Group} & \textbf{Equal} &
\textbf{One Factor} & \textbf{Pairwise} & \textbf{Group} & \textbf{Equal} \\
\midrule
Pick Place  & \textbf{69.5} & 68.6 & 62.0 & 56.1 & 73.2  &  \textbf{78.8} & 64.4 & 64.3 \\
Peg Insertion 
  & 22.4 & \textbf{26.0} & 22.2 & 20.2 & 41.9 & 38.1 & \textbf{45.3} & 28.1 \\
Pull Cube Tool 
  & 51.1 & \textbf{75.5} & 68.4 & 62.7 & 53.0 & 79.5 & \textbf{83.5} & 56.6 \\
\bottomrule
\end{tabular}
\vspace{4pt}
\label{table:xaxis}
\end{table*}

We provide ablation studies on different design choices of the curve construction.
Because the cost of \textbf{Pairwise} grows quadratically with $N$, we test it only on the tasks with visual factors, where $N=5$. In \cref{table:xaxis}, we find that in $K=20$ setting the performance drop of using \textbf{Group} compared to \textbf{Pairwise} is small, while \textbf{One Factor} is generally not good due to the small curve construction range.
At $K=100$, \textbf{Group} beats \textbf{Pairwise} except in the \emph{Pick Place} task. This is likely because \textbf{Group} heuristically filters out unrelated factor pairs based on human priors, whereas \textbf{Pairwise} becomes vulnerable to a single poorly-fitted curve among many. Furthermore, \textbf{Group} needs only 12 policies in this scenario, offering an order-of-magnitude lower cost while retaining the full performance advantage over the baselines.

\begin{table*}[h]
\centering
  \setlength{\tabcolsep}{6pt}
  \renewcommand{\arraystretch}{1.1}
    \caption{Ablation of data collection strategies. All the results are obtained using \textbf{Group} strategy for curve construction. We find that \textbf{Top} generally performs the best, in both simulation tasks and real world tasks.}

  \begin{tabular}{@{}lccc@{\hspace{16pt}}ccc@{}}
    \toprule
          & \multicolumn{3}{c}{$K{=}20$}
          & \multicolumn{3}{c}{$K{=}100$} \\
    \cmidrule(lr){2-4}\cmidrule(lr){5-7}
    \textbf{Task} &
    \textbf{Top} & \textbf{Top-Half} & \textbf{All} &
    \textbf{Top} & \textbf{Top-Half} & \textbf{All} \\
    \midrule
    Pick Place             & 62.0 & 62.0 & \textbf{67.5} &
                             62.0 & 62.0 & \textbf{70.2} \\
    Peg Insertion - Visual   & 22.2 & 22.2 & \textbf{29.5} &
                             \textbf{45.3} & 45.3 & 41.8 \\
    Peg Insertion - Spatial   & \textbf{45.5} & 40.8 & 39.0 &
                             \textbf{57.9} & 47.5 & 50.3 \\
    Pull Cube - Visual     & \textbf{68.4} & 68.4 & 55.3 &
                             \textbf{83.5} & 83.5 & 49.8 \\
    Pull Cube - Spatial   & \textbf{76.3} & 70.6 & 69.5 &
                             \textbf{83.4} & 75.3 & 79.1 \\
    Mouse in Drawer ($\pi_0$) & \textbf{58.3} & 33.3 & 31.3 & \textbf{66.7} & 29.2 & 33.3 \\
    \bottomrule
  \end{tabular}

  \label{table:pred}
\end{table*}
We also ablate the prediction strategies we use, see \cref{table:pred}. Among tasks with only visual factors (\(N=5\)), \textbf{Top} and \textbf{Top-Half} are the same as we pick $\left\lfloor \tfrac{N}{2} \right\rfloor$ factors for \textbf{Top-Half} strategy. \textbf{Top} delivers the best results in the last three tasks, where one factor group clearly dominates, matching the large gaps visible in their factored scaling curves (see~\cref{app:additional_curve}). 
However, in \emph{Pick Place}, factor importance is nearly uniform
(\cref{fig:sim-pick-place}); here the \textbf{All} rule prevails
because over-focusing on the top group hurts coverage. Hence, in practice, we can adopt a simple decision strategy:  If the curves show similar gains for all factors, use
\textbf{All}; if one factor group stands out, use \textbf{Top}.
Additional prediction-rule ablations under varying initial set sizes are
reported in~\cref{app:additional}. 

\subsection{How effective is FSC constructed with proxy metrics?}
\label{sec:emb_distance}


We additionally investigate the construction of factored scaling curves \emph{without} evaluating trained policies on hardware. 
Specifically, we explore the policy embedding similarity \cite{majumdar2025predictiveredteamingbreaking} as a proxy for the real-world success rate for guiding data collection. Given policy $\pi$ and two policy inputs $x_i$ and $x_j$, we define the embedding similarity $c_\pi$ to be the cosine similarity between the embeddings:
\begin{equation}
\label{Eqn:emb_similarity}
c_\pi(x_i,x_j)=\frac{\phi_\pi(x_i)\cdot\phi_\pi(x_j)}{||\phi_\pi(x_i)||\ ||\phi_\pi(x_j)||}
\end{equation}
where $\phi(\cdot)$ is the policy embedding, e.g., the output of the vision encoder. 

We define the training dataset $D_\text{train} = \{x_i\}_{i=1}^{N_\text{train}}$ that varies in environment factors, and an \emph{evaluation} (holdout) dataset $D_\text{eval} = \{x_i\}_{i=1}^{N_\text{eval}}$ collected in the target environment distribution.
\emph{Both datasets contain only the initial observation and thus collecting $D_\text{eval}$ does not require rolling out trajectories on hardware.}
We compute the embedding similarity between an input $x_i \in D_\text{eval}$ and $D_\text{train}$:
\begin{equation}
\label{Eqn:emb_point_to_set}
c_\pi(x_i, D_\text{train})= \max_{x_j \in D_\text{train}} c_\pi (x_i, x_j),
\end{equation}
which is maximized when there exist points in $ D_\text{train}$ that are similar to $x_i$. A generalization of ~\cref{Eqn:emb_point_to_set} is a $k$-nearest-neighbor variant, which averages the $k$ largest similarities between $x_i$ and $D_\text{train}$. After obtaining all $c_\pi(x_i, D_\text{train})$, we normalize them to $[0,1]$. 
Then, we define the \textbf{policy embedding similarity} $\bar{c}_\pi$ as the embedding similarity between the two datasets $D_\text{train}$ and $D_\text{eval}$ averaged over instances in $D_\text{eval}$:
\begin{equation}
\label{Eqn:emb_distance}
\bar{c}_\pi=\sum_{x_i\in D_\text{eval}}\frac{c(x_i,D_\text{train})}{|D_\text{eval}|}.
\end{equation}
Intuitively, higher policy embedding similarity $\bar{c}_\pi$, indicating consistent behavior of the policy between environments where the data is collected and those where the policy is evaluated, should correspond to higher performance at the target environments. After obtaining the embedding similarity $\bar{c}_\pi$ for each policy $\pi$, we construct the factored scaling curve with it and use the \textbf{Top} strategy to collect data, following \cref{alg:fsc} and \cref{alg:gdc}.

\begin{table*}[h]
  \centering
  \small
  \setlength{\tabcolsep}{3pt}    
  \renewcommand{\arraystretch}{1.05}
  \caption{Success rates $(\%)$ on simulation tasks when guiding data collection with factored scaling curves built from embedding similarity of diffusion policy (\textbf{FSC-Proxy}). For \emph{Peg Insertion} and \emph{Pull Cube Tool}, we show results with spatial factors. For both small ($K=20$) and large ($K=100$) data‑collection budgets, \textbf{FSC-Proxy} matches or surpasses the original \textbf{FSC} and consistently outperforms the baselines. }
  \label{tab:dp-embedding}
  \begin{tabular}{
          l         
          *{3}{c}   
          @{\hspace{6pt}}  
          *{3}{c}   
        }
    \toprule
    & \multicolumn{3}{c}{$K=20$} & \multicolumn{3}{c}{$K=100$} \\
    \cmidrule(lr){2-4}\cmidrule(lr){5-7}
    Method & Pick Place & Peg Insertion & Pull Cube Tool & Pick Place & Peg Insertion & Pull Cube Tool \\
    \midrule
    Equal                & 56.1 & 43.8 & 57.7 & 64.3 & 49.5 & 78.5 \\
    Greedy               & 58.7 & 42.0 & 73.3 & 65.9 & 52.7 & 62.5 \\
    Re-Mix               & 61.6 & 31.7 & 50.5 & 64.7 & 44.1 & 64.5 \\    
    \midrule
    \textbf{FSC-Proxy}    & \textbf{70.9} & 45.2 & 73.5 & \textbf{74.1} & 53.3 & 73.4 \\
    \textbf{FSC}                 & 62.0 & \textbf{45.5} & \textbf{76.3} & 64.4 & \textbf{57.9} & \textbf{83.4} \\
    \bottomrule
  \end{tabular}
\end{table*}

We report results for Diffusion Policy (DP) \cite{chi2023diffusionpolicy} and $\pi_0$ \cite{black2024pi0visionlanguageactionflowmodel}.
For DP, we use the output feature from the vision encoder (ResNet‑18~\cite{he2015deepresiduallearningimage}) as our embedding $\phi(\cdot)$. 
We tabulate results for DP in \cref{tab:dp-embedding}, and show the result for real-world \emph{Pick Place} task in \cref{fig:real_exp}. We use $k=1$ for \textbf{FSC-Proxy} for the $k$-nearest-neighbor step, and ablate other choices of $k$ in~\cref{app:extra_emb_result}. Generally, we find that \textbf{FSC-Proxy} achieves performance comparable to \textbf{FSC}, sometimes even surpassing it, while consistently outperforming the baseline methods. 
Our results provide preliminary evidence on the effectiveness of using embedding similarity as a \textbf{surrogate metric for guiding data collection} in place of success rates from expensive real-world evaluations.

For $\pi_0$, we define $\phi(\cdot)$ to be the attention weights from the final denoising step of the flow-matching-based action expert \cite{black2024pi0visionlanguageactionflowmodel}. We take the mean weight over each attention head and action token so that the embedding has the same size as the VLM sequence length. We define $D_\text{train}$ and $D_\text{eval}$ in the same way as DP. As shown in \cref{fig:real_exp}, \textbf{FSC-Proxy} successfully prioritizes the same factor for data collection as \ACRO{} for the \emph{Fold Towel - Spatial} and \emph{Mouse in Drawer} task, achieving the highest success rate. This further shows that embedding similarity is an effective surrogate metric for guiding data collection for pre-trained VLA models. We additionally visualize the correlations between embedding similarity and real success rate and ablate other embedding choices in \cref{app:extra_emb_result}.

\section{Conclusions}
\label{sec:conclusion}

We propose \emph{Factored Scaling Curves}, which quantify how a policy’s performance improves as additional data is collected involving different factor variations. We show that factored scaling curves can be reliably extrapolated to make predictions about how policy performance evolves if we collect more data for the factor. We leverage this property to propose a principled way to guide data collection, where we decide priority of the factors to collect data for based on the slopes of their respective factored scaling curve. We empirically study different ways of constructing the factored scaling curve, and propose varying factors in groups to strike a strong balance between evaluation cost and performance. We also study different ways of allocating the data budget, and find that allocating the entire budget to the most promising factor(s) performs best. We study a wide range of simulation tasks and real-world tasks, including ones where we train from scratch and fine-tune a pre-trained VLA. Overall, our method can achieve up to 26\% success rate improvement compared to state-of-the-art data collection methods.

\paragraph{Limitations and Future Work.}We discuss the limitations of \ACRO{} and outline future work to address them. Although we have shown that embedding-space similarity provides a strong proxy for real-world success—yielding curves that closely track and effectively guide data collection—curves built with the actual success rate remain marginally more predictive. This superior fidelity comes at a cost: obtaining real-world success rates demands on-hardware evaluation and thus substantial human effort (roughly 10–20 trials per policy–factor pair). Future research should therefore focus on further boosting the reliability of purely offline metrics—such as embedding-space distance or simulation success—so that practitioners can confidently construct scaling curves without incurring expensive physical evaluations. In the meantime, users can choose between lower-cost embedding metrics and higher-accuracy real success rates, depending on their resource constraints and precision requirements.

Second, as \ACRO{} requires extrapolating the existing curve, the prediction at large $K$ (large data budget) can be less precise as shown in \cref{table:app_diff_dataset_spatial}. For such settings, a more adaptive version of \ACRO{} might be useful as the practitioner collects additional data and re-evaluates the policy before deciding on the next factors to collect data with.

Lastly, in this work we primarily consider settings where we use a pre-trained policy or collect data from scratch. It would be interesting to extend \ACRO{} to the retrieval setting \cite{du2023behavior, di2024effectiveness} where a large dataset is given and factored scaling curves can help determine which factors of data are more useful to policy performance. \ACRO{} may also be applied to pre-training in this setting.
\acknowledgments{%
  This work was partially supported by the NSF CAREER Award~\#2044149,
  the Office of Naval Research~(N00014-23-1-2148),
  and a Sloan Fellowship.}
\bibliography{main}  

\clearpage
\appendix
\section{Additional Results}
\label{app:additional}

\subsection{Algorithms}
\label{sec:algorithms}
We present the construction of factored scaling curves and the subsequent data collection strategies. We provide pseudocode for curve construction and data collection strategy in the \textbf{Group} setting of pairs of factors. For this setting, curve construction requires the following inputs. First, a policy parametrization \(\pi\) denotes the policy (e.g., diffusion policy~\cite{chi2023diffusionpolicy} and \(\pi_0\)~\cite{black2024pi0visionlanguageactionflowmodel}) trained on varying amounts of data as a part of scaling curve construction. Second, a set of training demonstrations \(\mathcal{D}\) to guide further data collection. Third, a set of \emph{factor combinations \(\mathcal{F}_{\text{group}}\)} specified by the \textbf{Group} setting, which divides \(N\) factors into \(\lceil N/2\rceil\) factor pairs. We construct a factored scaling curve for each factor combination. Finally, we require a metric \(S\) to evaluate the policy on a fixed set of evaluation environments.
In addition to these inputs, we set a hyperparameter \(m\) which sets the number of points used to construct the scaling curve. 
\begin{algorithm}
\caption{Factored Scaling Curves (Construction)}\label{alg:fsc}
\begin{algorithmic}[1]
\Require Policy parametrization \(\pi\), demonstrations \(\mathcal{D}\), factor combinations \(\mathcal{F}_{\text{group}}\), metric \(S\), hyperparameter \(m\) 
\Ensure A set of factored scaling curves \(\{\hat{\Phi}_{ij} \,\vert\, \{f_i, f_j\} \in \mathcal{F}_{\text{group}}\}\), one for each factor combination.
\For{each factor combination $\{f_i, f_j\} \in \mathcal{F}_{\text{group}}$} 
\State Factor combination dataset sizes for training \(\mathcal{N} = \{\frac{\lvert \D{ij}\rvert(i-1)}{m-1} \,\vert \, i \in \{1,\ldots, m\}\}\)
\For{\(k \in \mathcal{N}\)}
\State Assemble training dataset \(\D{ij}^k \coloneqq (\mathcal{D} \setminus \mathcal{D}_{ij}) \cup  \dD{ij}^k\)
\State Train policy \(\pi(\D{ij}^k)\)
\State Record policy performance \(S(\pi(\D{ij}^k))\)
\EndFor
\State Construct \(\hat{\Phi}_{ij}\) by fitting points \(\{(k, S(\pi(\D{ij}^k)))\}_{k \in \mathcal{N}}\) according to a power-law (\cref{eq:fsc_construction}). 
\EndFor
\end{algorithmic}
\end{algorithm}

Following~\cref{alg:fsc}, we can use the constructed factor scaling curves to determine a data collection strategy for some data budget \(K\). We consider three strategies for splitting the data budget amongst factor combinations: \textbf{Top}, \textbf{Top-Half}, and \textbf{All}.
\begin{algorithm}
\caption{Data collection guided by Factored Scaling Curves}\label{alg:gdc}
\begin{algorithmic}[1]
\Require Factored scaling curves \(\{\hat{\Phi}_{ij}\}\), factor combinations \(\mathcal{F}_{\text{group}}\), factors \(\mathcal{F}\), data budget \(K\)
\Ensure Recommendation of additional dataset size \(\lvert \Delta \D{i}\rvert\) for each factor \(f_i\)
\State Initialize \(\lvert \Delta \D{i}\rvert = 0\) for each factor \(f_i \in \mathcal{F}\)
\For{each factor combination $\{f_i, f_j\} \in \mathcal{F}_{\text{group}}$} 
\State \(P^K_{ij}\leftarrow\) Approximate the slope of \textbf{FSC} \(\hat{\Phi}_{ij}\) using~\cref{Eqn:Pij}
\EndFor
\State Rank all pairs in \(\mathcal{F}_{\text{group}}\) by slope \(P^K_{ij}\) in descending order
\State \(\mathcal{G}_{inc} = set()\) \Comment{To store factor combinations selected for data allocation}
\If{strategy is \textbf{Top}}
\State \(\mathcal{G}_{inc} \leftarrow \{(i^*, j^*)\}\), where \(P^K_{i^*j^*} = \max_{ij} P^K_{ij}\) 
\ElsIf{strategy is \textbf{Top-Half}}
\State \(\mathcal{G}_{inc} \leftarrow\) Set of top \(\lceil \lvert \mathcal{F}_{\text{group}} \rvert / 2 \rceil\) pairs
\Else \Comment{strategy is \textbf{All}}
  \State \(\mathcal{G}_{\text{inc}} \leftarrow \mathcal{F}_{\text{group}}\)
\EndIf
\For{each factor combination $\{f_i, f_j\} \in \mathcal{G}_{inc}$} 
\State Allocate \(\lvert \Delta \D{ij}\rvert\) proportionally using~\cref{eq:data_alloc_group}.
\State \(\lvert \Delta \D{i}\rvert \leftarrow \lvert \Delta \D{i}\rvert + \lvert \Delta \D{ij}\rvert \frac{\lvert \D{i} \rvert}{\lvert \D{ij} \rvert}\) \Comment{Divide pairwise allocation in half}
\EndFor
\end{algorithmic}
\end{algorithm}
\clearpage

\subsection{Data Collection Strategies for Factor Combinations}
\label{app:data_collect}
The two-factor analog to the factor dataset is denoted by \(\D{ij} = \D{i} \cup \D{j}\) for factors \(f_i\) and \(f_j\), and \(\D{ij}^{n}\) follows as \(\mathcal{D}^{n}_{ij} \coloneqq (\mathcal{D} \setminus \mathcal{D}_{ij}) \cup  (\dD{i}^{n_i} \cup \dD{j}^{n_j})\), where \(n_i + n_j = n\) and are proportional to the sizes of \(\D{i}\) and \(\D{j}\). We choose \(\lvert \D{i} \rvert = \frac{\lvert \mathcal{D} \rvert}{N}\), for all \(i\), which forms a uniform prior on factor importance. The combination of \(f_i\) and \(f_j\) is denoted \(f_{ij}\), and the scaling curve is referred by \(\hat{\Phi}_{ij}\). 

Recall that \(K\) is the total budget allocated for new demonstrations. We present the data collection strategy for factor combinations (i.e., \textbf{Pairwise} and \textbf{Group}), which covers the three methods presented in~\cref{sec:fsc}. For two-factor pairs, we let $\mathcal{G}_2$ denote the set of all index pairs. For each factor combination, the predicted policy performance after adding \(K\) demonstrations is \(\hat{\curve}_{ij}\!\bigl(\lvert\mathcal{D}_{ij}\rvert+K\bigr)\). We coarsely approximate the slope of the scaling curve as
\begin{equation}
\label{Eqn:Pij}
P^K_{ij}\;:=\;\frac{\hat{\curve}_{ij}\!\bigl(\lvert\mathcal{D}_{ij}\rvert+K\bigr) - \hat{\curve}_{ij}\!\bigl(\lvert\mathcal{D}_{ij}\rvert\bigr)}{K}.
\end{equation}
Based on \cref{Eqn:Pij} we consider three strategies that vary in index inclusion set $\mathcal{G}_{inc}$: {\bf (1) Top:} Identify the factor combination $f_{i^*j^*}$ with fastest predicted performance gain \(P^K_{i^*j^*}\) and set $\mathcal{G}_{inc}=\{(i^*,j^*)\}$; {\bf (2) Top-Half:} Identify the top half of the factor combinations according to \(P^K_{ij}\) and set $\mathcal{G}_{inc}$ to contain half of the two-factor indices; {\bf (3) All:} Spread the budget over \emph{all} factor combinations and set $\mathcal{G}_{inc}=\mathcal{G}_2$. New demonstrations are allocated by:
\begin{equation}
    \label{eq:data_alloc}
    \lvert\Delta\mathcal{D}_{i}\rvert\;=\; \frac{\sum_{j}P^K_{ij}}{2\sum_{(i', j')} P^K_{i'j'}}\;K, 
\end{equation}
and $\lvert \Delta \D{i} \rvert = 0$ if no pair in $\mathcal{G}_{inc}$ contains index $i$. We evaluate each of these strategies in the subsequent experiments.


\subsection{Further Analysis on Embedding Similarity for Guiding Data Collection}
\label{app:extra_emb_result}
\begin{figure}[h]
    \centering
    \includegraphics[width=\linewidth]{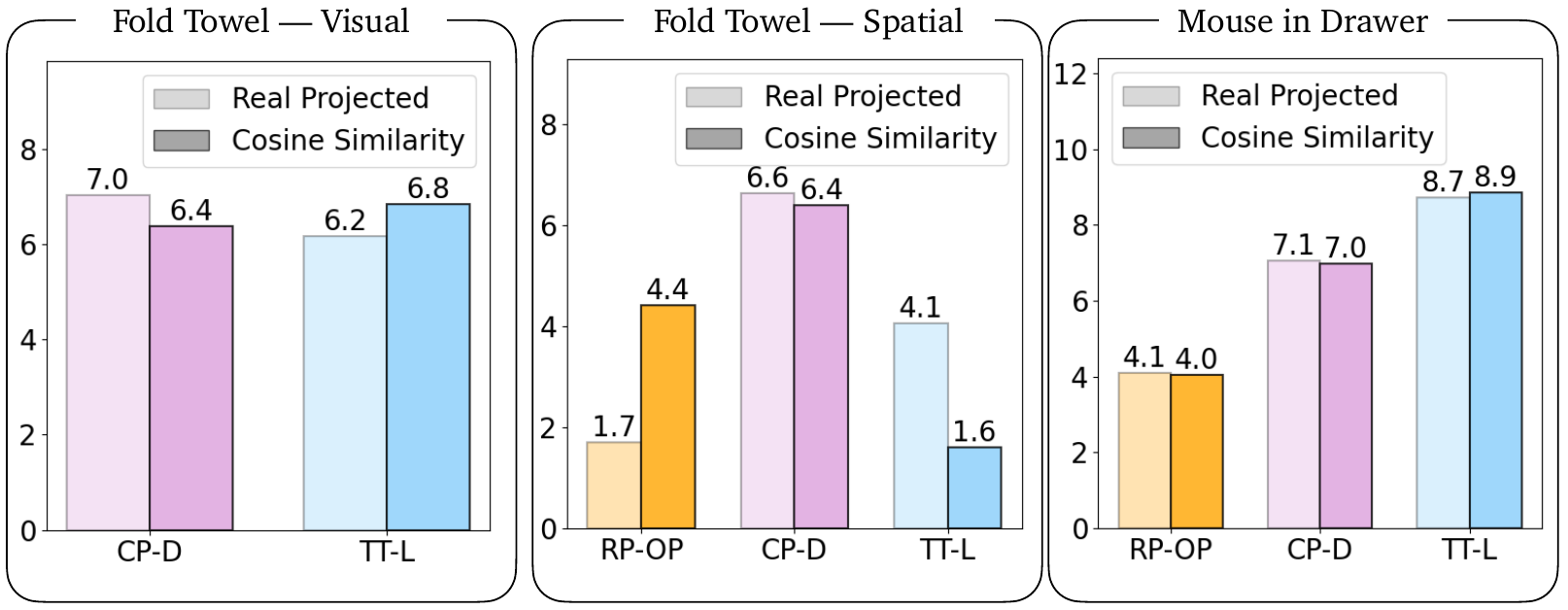}
    \caption{Expected improvement for $\pi_0$ on three task settings using the \textbf{Attention Weights} from the last denoising step: Camera Pose -- Distractor (CP-D), Table Texture -- Lighting (TT-L), and Robot Pose -- Object Pose (RP-OP). Cosine Similarity projections are normalized to have the same expected value as Expected Improvement. Cosine similarity predicts the top-ranked expected improvement for \textit{Fold Towel} (CP-D) and \textit{Mouse in Drawer} (TT-L).}
    \label{fig:proj-attn}
\end{figure}

\begin{figure}[h]
    \centering
    \includegraphics[width=\linewidth]{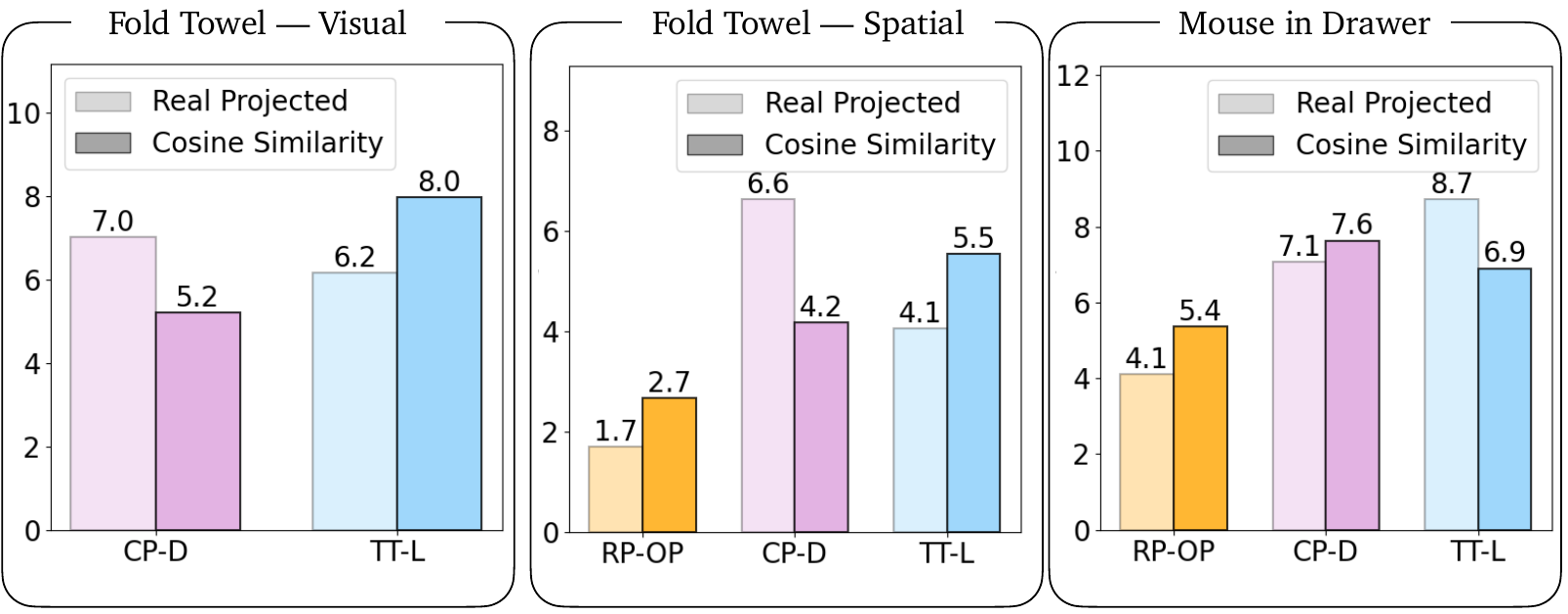}
    \caption{Expected improvement for $\pi_0$ on three task settings using the \textbf{Latent Action} from the first denoising step: Camera Pose -- Distractor (CP-D), Table Texture -- Lighting (TT-L), and Robot Pose -- Object Pose (RP-OP). Cosine Similarity projections are normalized to have the same expected value as Expected Improvement. Cosine similarity predicts the bottom-ranked expected improvement for \textit{Fold Towel} (RP-OP) and \textit{Mouse in Drawer} (RP-OP).}
    \label{fig:proj-embed}
\end{figure}

In addition to attention weights, we further investigate another embedding option $\phi(\cdot)$ for $\pi_0$ \cite{black2024pi0visionlanguageactionflowmodel}: the latent action vector after the first denoising step. We analyze the correlation between different embedding options and real success rate.
We report the results for \textbf{Attention Weights} in ~\cref{fig:proj-attn} and summarize two important findings here. Attention weights successfully predict the \textit{first ranked} factor in Fold Towel -- Spatial and Mouse in Drawer, offering some evidence that they may be used as a proxy for the \textbf{Top} data collection strategy---for example, if real data is scarce---when factors can be clearly differentiated.  We additionally conclude that while the attention weights may not always report the correct \textit{ranking} of factors (for example, the two factors of Fold Towel -- Visual and the lesser two factors of Fold Towel -- Spatial), the relative \textit{ratio} of factors remains accurate across all experiments, which indicates a close match to the data ratio predicted by the \textbf{All} strategy. In ~\cref{fig:proj-embed}, we report results for the \textbf{Latent Action} and conclude that it may be used to filter out the \textit{last ranked} factor in Fold Towel -- Spatial and Mouse in Drawer. We observe a similar trend in the ratio between factors, which suggests using the \textbf{All} strategy.

We then ablate the different choices of $k$, where $k$ denotes the value used in the $k$‑nearest‑neighbors step. As shown in \cref{tab:emb_k_ablation}, performance is similar across different $k$ values, with \textbf{FSC-Proxy $\boldsymbol{(k=1)}$} performing slightly better in the $K=20$ setting and \textbf{FSC-Proxy $\boldsymbol{(k=5)}$} performing slightly better in the $K=100$ setting. This indicates that \textbf{FSC-Proxy} is not sensitive to the hyper‑parameter $k$, and that $k=1$ or $k=5$ are generally good choices depending on the dataset size. 
\begin{table*}[h]
  \centering
  \small
  \setlength{\tabcolsep}{3pt}    
  \renewcommand{\arraystretch}{1.05}
  \caption{Ablations on different choices of $k$ for \textbf{FSC-Proxy} used for $k$‑nearest‑neighbor filtering. For \emph{Peg Insertion} and \emph{Pull Cube Tool}, we show results with spatial factors. Overall, \textbf{FSC-Proxy} exhibits comparable performance under different $k$ in most settings, indicating that it is insensitive to the choice of hyperparameter $k$.}
  \label{tab:emb_k_ablation}
  \begin{tabular}{
          l         
          *{3}{c}   
          @{\hspace{6pt}}  
          *{3}{c}   
        }
    \toprule
    & \multicolumn{3}{c}{$K=20$} & \multicolumn{3}{c}{$K=100$} \\
    \cmidrule(lr){2-4}\cmidrule(lr){5-7}
    Method & Pick Place & Peg Insertion & Pull Cube Tool & Pick Place & Peg Insertion & Pull Cube Tool \\
    \midrule
    \textbf{FSC-Proxy $\boldsymbol{(k=1)}$}    & \textbf{70.9} & 45.2 & \textbf{73.5} & \textbf{74.1} & 53.3 & 73.4 \\
    \textbf{FSC-Proxy $\boldsymbol{(k=5)}$}    & 68.6 & \textbf{45.5} & 41.7 & 73.2 & 55.0 & \textbf{86.6} \\
    \textbf{FSC-Proxy $\boldsymbol{(k=10)}$}   & 69.9 & 44.1 & 66.5 & 71.5 & \textbf{56.2} & 79.2 \\
    \bottomrule
  \end{tabular}
\end{table*}

\subsection{Ablating Different Initial Dataset Size and Prediction Horizon}

We further investigate whether \ACRO{} maintains strong performance under different initial‑dataset sizes. In \cref{table:app_diff_dataset_visual}, we show that when the initial dataset contains 300 demonstrations—double the 150‑demonstration setting reported in \cref{table:simulation}—our method attains performance comparable with the baseline. This result is unsurprising, as task performance appears to have already saturated in this data regime.

\begin{table*}[h]
\centering
\renewcommand{\arraystretch}{1.05}
\caption{Ablation on different initial dataset size on the \emph{Peg Insertion - Visual} task. Initial dataset contains 300 demos.}
\begin{tabular}{
  l
  c
  c
  c
}
\toprule
 &\textbf{Top} & \textbf{All} & \textbf{Equal} \\
\midrule
\addlinespace
$K=20$ & 58.4 & \textbf{64.9}  & 64.2 \\
$K=100$ & 49.3 & \textbf{53.4} & 52.5 \\
\bottomrule
\end{tabular}
\vspace{4pt}

\label{table:app_diff_dataset_visual}
\end{table*}

In \cref{table:app_diff_dataset_spatial}, we further examine how \ACRO{} performs under different initial dataset size in another task, as well as how accurately \ACRO{} predicts policy performance over an even longer horizon. We evaluate settings with up to $K=500$ additional demonstrations, starting from an initial dataset of 480 demonstrations (as opposed to the 240‑demonstration setting used in the main results). In the low‑data regime ($K=20$), \textbf{Top} achieves the best performance. As the data budget increases, \textbf{All} becomes superior, likely because the factors emphasized by \textbf{Top} have already saturated, while \textbf{All} distributes additional demonstrations across all factors according to their estimated importance instead of exploiting only the top combination. Interestingly, at $K=500$ the performance of \textbf{All} falls by roughly $10\%$. We hypothesize that this drop stems from performance saturation in this regime, compounded by substantial evaluation noise—particularly salient because the peg‑insertion task demands high precision.

\begin{table*}[h]
\centering
\renewcommand{\arraystretch}{1.05}
\caption{Ablation on different initial dataset size on the \emph{Peg Insertion - Spatial} task. Initial dataset contains 480 demos.}
\begin{tabular}{
  l
  c
  c
  c
  c
}
\toprule
 &\textbf{Top} & \textbf{Top-Half} & \textbf{All} & \textbf{Equal} \\
\midrule
\addlinespace
$K=20$ & \textbf{67.1}  & 64.1  & 65.6 & 68.5 \\
$K=40$ &  66.2 &  63.4 & \textbf{68.9} & 63.4 \\
$K=100$ &  62.3 &  62.8 & \textbf{72.4} & 56.0 \\
$K=250$ &  55.4 & 55.3  & \textbf{69.1} & 61.4 \\
$K=500$ &  56.5 &  48.4 & 59.4 & \textbf{63.0} \\
\bottomrule
\end{tabular}
\vspace{4pt}

\label{table:app_diff_dataset_spatial}
\end{table*}

\subsection{Additional Curve Visualization}
\label{app:additional_curve}

In this section, we visualize the factored scaling curves for all experiments.

\begin{figure}[H]
    \centering
    \includegraphics[width=\linewidth]{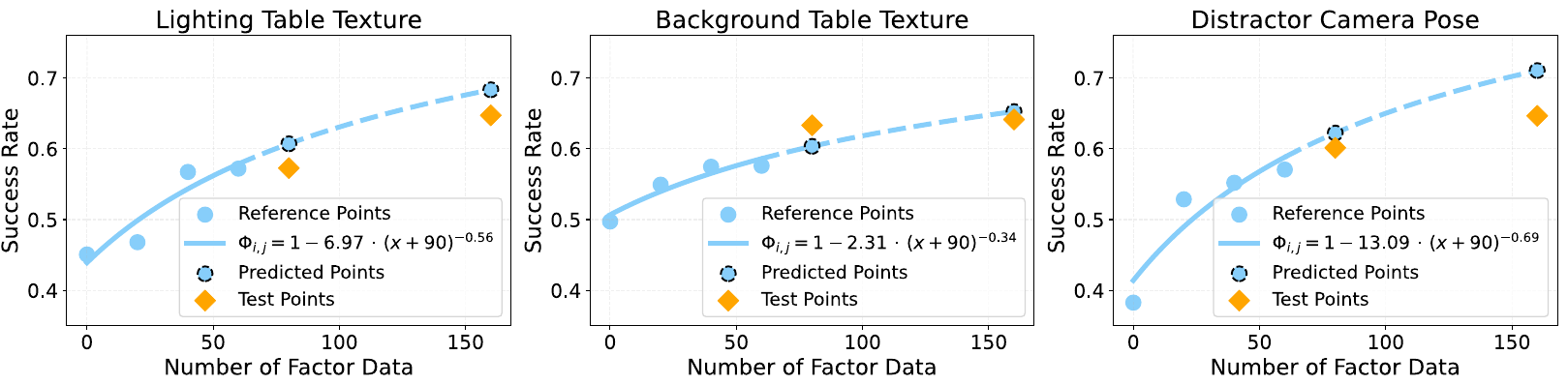}
    \caption{Factored scaling curves for the simulation \emph{Pick Place} task.}
    \label{fig:sim-pick-place}
\end{figure}

\begin{figure}[h]
    \centering
    \includegraphics[width=\linewidth]{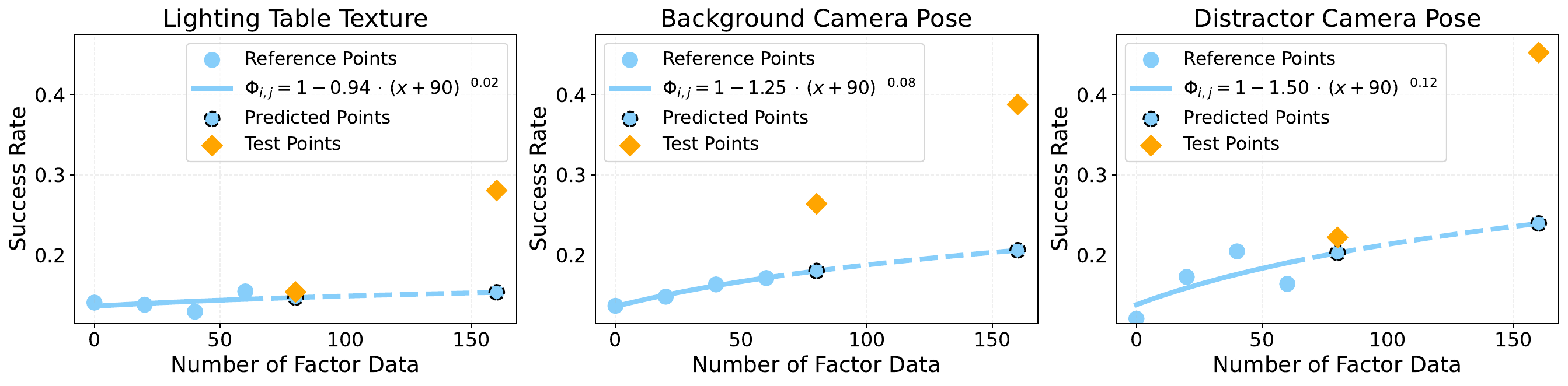}
    \caption{Factored scaling curves for the simulation \emph{Peg Insertion - Visual} task.}
    \label{fig:sim-peg-insertion-visual}
\end{figure}
\vspace{10pt}
\begin{figure}[h]
    \centering
    \includegraphics[width=\linewidth]{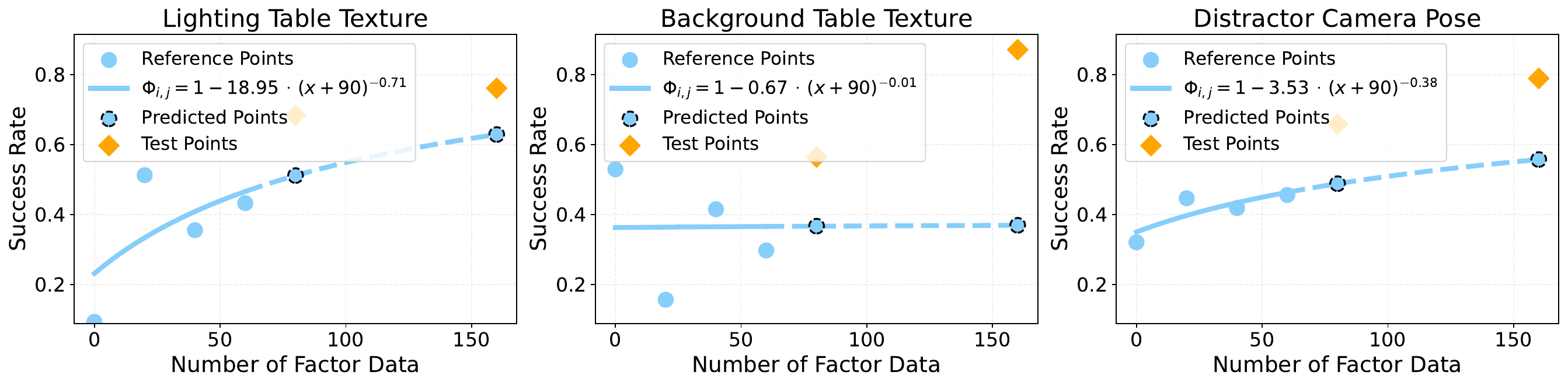}
    \caption{Factored scaling curves for the simulation \emph{Pull Cube Tool - Visual} task.}
    \label{fig:sim-pull-visual}
\end{figure}

\begin{figure}[htbp]
    \centering
    \includegraphics[width=\linewidth]{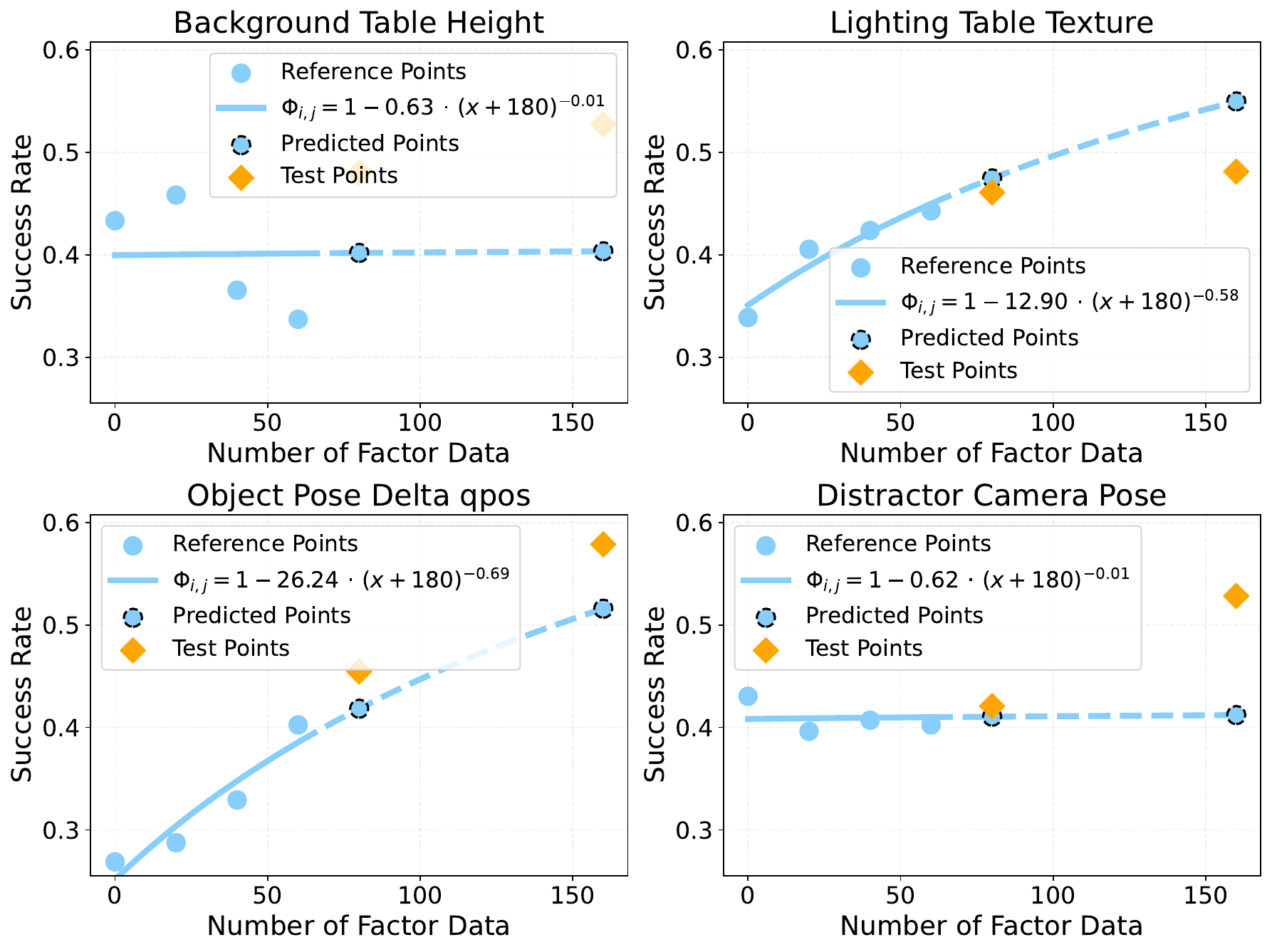}
    \caption{Factored scaling curves for the simulation \emph{Peg Insertion - Spatial} task.}
    \label{fig:sim-peg-insertion-spatial}
\end{figure}

\clearpage

\begin{figure}[htbp]
    \centering
    \includegraphics[width=\linewidth]{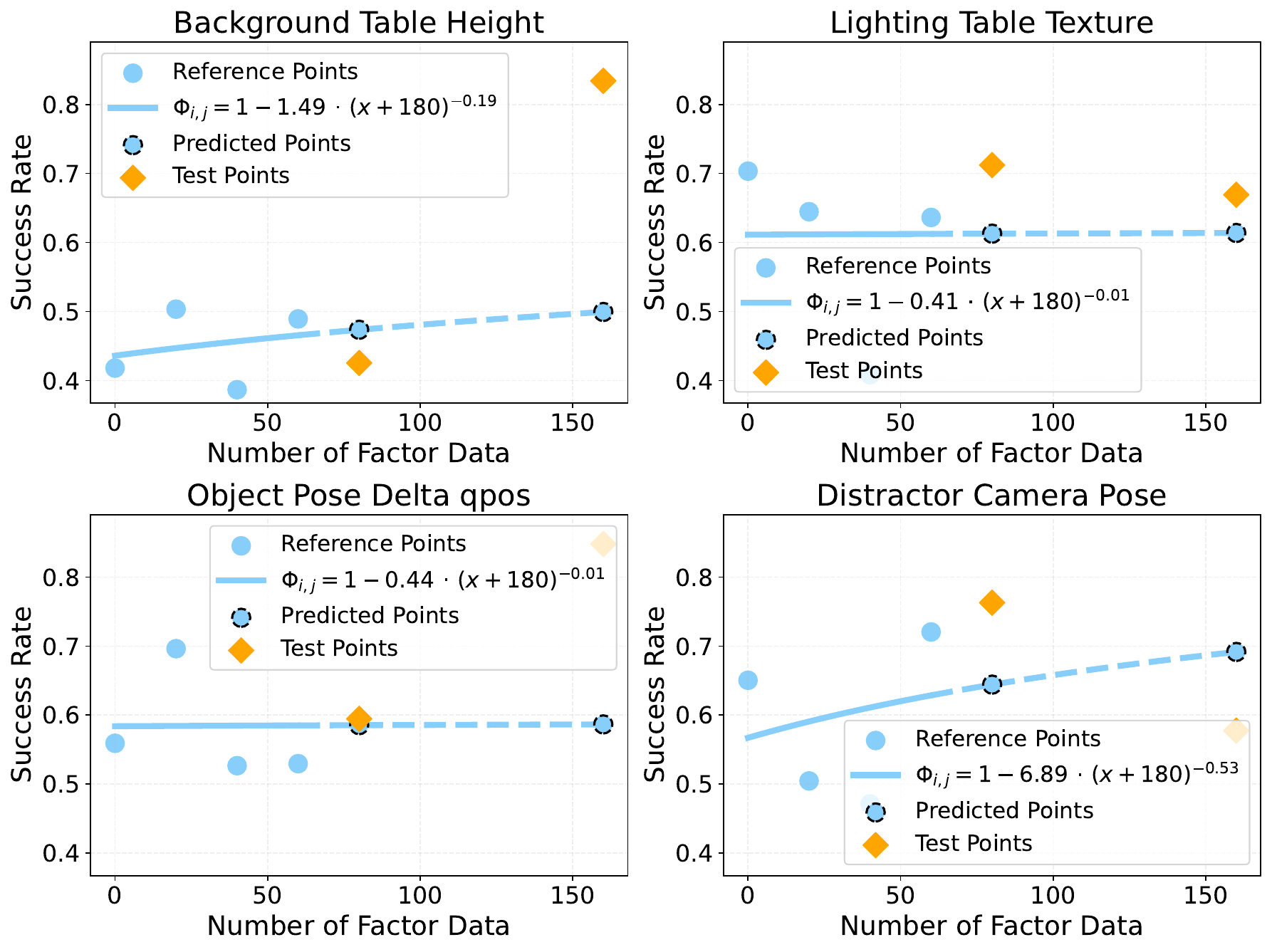}
    \caption{Factored scaling curves for the simulation \emph{Pull Cube Tool - Spatial} task.}
    \label{fig:sim-pull-spatial}
\end{figure}


\begin{figure}[h]
    \centering
    \includegraphics[width=\linewidth]{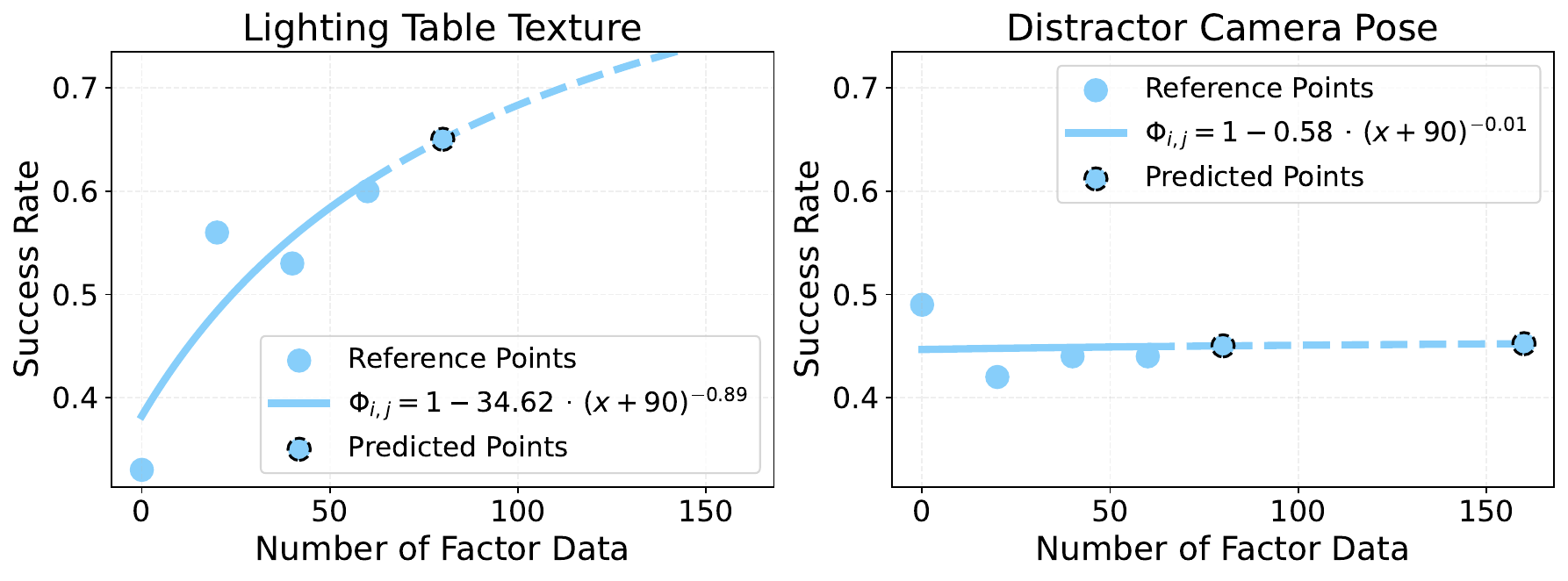}
    \caption{Factored scaling curves for the real \emph{Pick Place} task. For real world tasks, we do not obtain the ground truth test points for visualization.}
    \label{fig:sim-pull-spatial}
\end{figure}

\begin{figure*}[t]                         
  \centering
  \begin{subfigure}[b]{0.32\textwidth}     
  \centering
    \includegraphics[width=\linewidth]{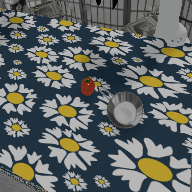}
    \caption{Pick Place}
    \label{fig:fsc-a}
  \end{subfigure}
  \hfill                                
  \begin{subfigure}[b]{0.32\textwidth}
  \centering
    \includegraphics[width=\linewidth]{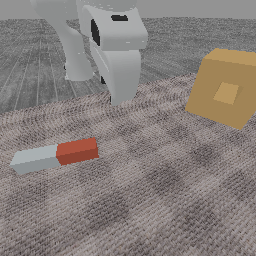}
    \caption{Peg Insertion}
    \label{fig:fsc-b}
  \end{subfigure}
  \hfill 
  \begin{subfigure}[b]{0.32\textwidth}
  \centering
    \includegraphics[width=\linewidth]{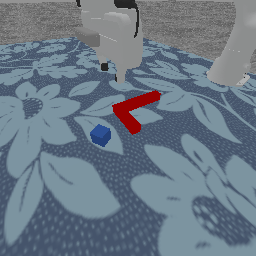}
    \caption{Pull Cube Tool}
    \label{fig:fsc-c}
  \end{subfigure}
  \caption{Illustrations of simulation tasks.}
  \label{fig:sim_tasks}
\end{figure*}

\section{Simulation Experiments}
\label{app:sim_exp_setup}
\newlength{\subfigheight}
\setlength{\subfigheight}{0.28\textheight}

\begin{figure*}[htbp]
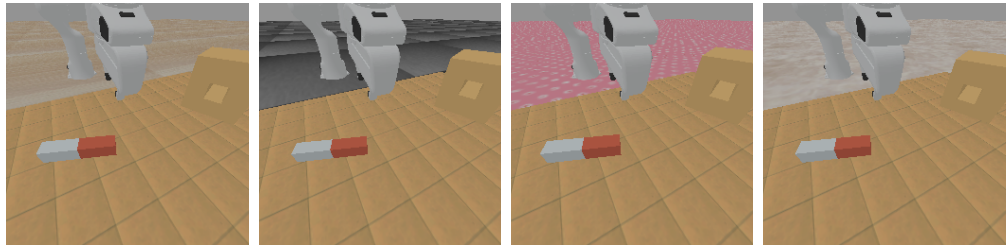
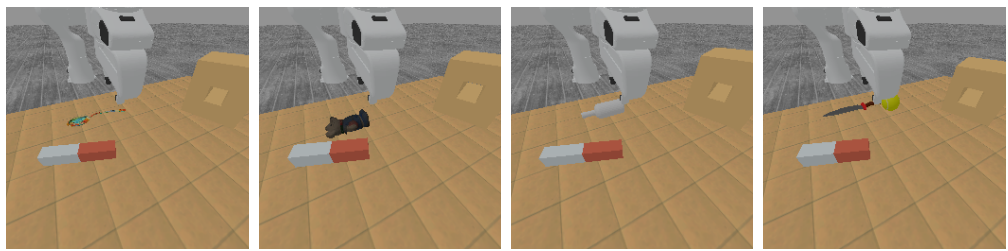
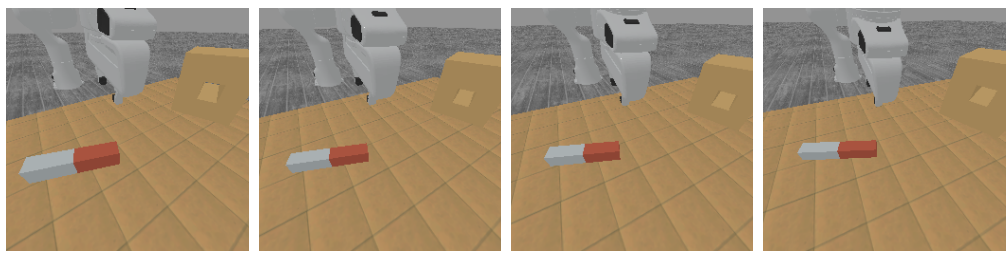
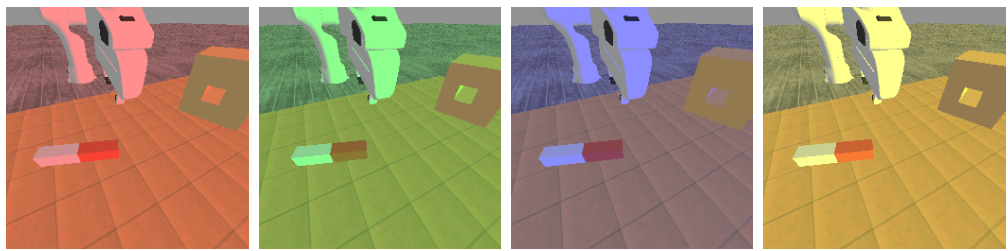
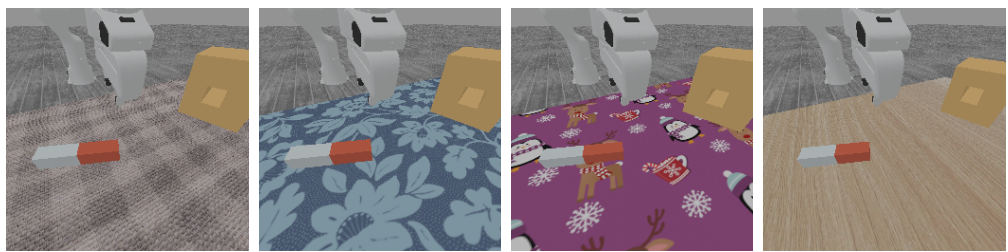

\centering
\newcommand{\simvar}[2]{\includegraphics[width=0.23\textwidth]{figures/sim_variation/#1_#2.png}}
\begin{subfigure}[t]{\textwidth}
    \centering
    \simvar{background}{0}\hspace{4pt}%
    \simvar{background}{1}\hspace{4pt}%
    \simvar{background}{2}\hspace{4pt}%
    \simvar{background}{3}
    \caption{Background}
    \vspace{5pt}
\end{subfigure}

\begin{subfigure}[t]{\textwidth}
    \centering
    \simvar{distractor}{0}\hspace{4pt}%
    \simvar{distractor}{1}\hspace{4pt}%
    \simvar{distractor}{2}\hspace{4pt}%
    \simvar{distractor}{3}
    \caption{Distractor Objects}
    \vspace{5pt}
\end{subfigure}

\begin{subfigure}[t]{\textwidth}
    \centering
    \simvar{camera_pose}{0}\hspace{4pt}%
    \simvar{camera_pose}{1}\hspace{4pt}%
    \simvar{camera_pose}{2}\hspace{4pt}%
    \simvar{camera_pose}{3}
    \caption{Camera Pose}
    \vspace{5pt}
\end{subfigure}

\begin{subfigure}[t]{\textwidth}
    \centering
    \simvar{directional}{0}\hspace{4pt}%
    \simvar{directional}{1}\hspace{4pt}%
    \simvar{directional}{2}\hspace{4pt}%
    \simvar{directional}{3}
    \caption{Lighting}
    \vspace{5pt}
\end{subfigure}

\begin{subfigure}[t]{\textwidth}
    \centering
    \simvar{table_texture}{0}\hspace{4pt}%
    \simvar{table_texture}{1}\hspace{4pt}%
    \simvar{table_texture}{2}\hspace{4pt}%
    \simvar{table_texture}{3}
    \caption{Table Texture}
\end{subfigure}

\caption{\small{Visualization of simulation environment visual factor variations.}}
\label{fig:sim_variation_factors}
\end{figure*}

All experiments are done in Maniskill3 \cite{taomaniskill3} on a Franka Panda robot. 

\subsection{Task and Factor Description}
\label{app:sim_task_description}

We visualize all simulation tasks in \cref{fig:sim_tasks}. To collect training data, we sample continuous‑factor values according to \cref{tab:sim_variation}. Note that robot pose and table height are varied only in experiments that involve spatial factors. For tasks with two cameras, we only vary the pose of the third-person view camera. The object‑pose range shown in \cref{tab:sim_variation} is used for all data except the object‑pose‑variation subset, for which we extend the range by an additional $25\%$.

For table‑texture and background variations, we draw four instances from a fixed texture dataset. We also prepare four sets of distractors for the distractor‑factor variation, each set containing two objects (e.g., eggplant, cup, cucumber). All visual factors are illustrated in \cref{fig:sim_variation_factors}.

\setlength{\tabcolsep}{.1cm}
\begin{table}[h]
\centering
\renewcommand{\arraystretch}{1.3}
\caption{Range for each continuous factor in meters for simulation tasks. 
}
\begin{tabular}{lllll}
    \toprule
    Factor & Parameters & Pick Place & Peg Insertion & Pull Cube Tool \\
    \midrule
    \multirow{3}{*}{Manipulated object pose} & X-position & $[-0.2, 0.2]$ & $[-0.04, 0.04]$ & $[-0.04, 0.04]$ \\
                              & Y-position &  $[-0.2, 0.2]$ & $[-0.04, 0.04]$ & $[-0.08, 0.08]$ \\
                              & Yaw & - & $[-0.13,0.13]$ & - \\
    \midrule
    \multirow{3}{*}{Goal object pose} & X-position & $[-0.15, 0.15]$ & $[-0.04, 0.04]$ & $[-0.04, 0.04]$ \\
                              & Y-position & $[-0.2, 0.2]$ & $[-0.04, 0.04]$ & $[-0.08, 0.08]$ \\
                              & Yaw & - & $[-0.13,0.13]$ & $[-0.13,0.13]$ \\

    \midrule
    \multirow{3}{*}{Camera position} & Eye-X & $[-0.05, 0.05]$ & $[-0.025, 0.025]$ & $[-0.05, 0.05]$ \\
                              & Eye-Y & $[-0.1,0.1]$ & $[-0.025, 0.025]$ & $[-0.05, 0.05]$ \\
                              & Eye-Z & $[-0.1,0.1]$ & $[-0.025, 0.025]$ & $[-0.05, 0.05]$ \\
    \midrule
    \multirow{1}{*}{Robot pose} & Initial joint angles & - & $[-0.015, 0.015]$ & $[-0.01, 0.01]$ \\
    \midrule
    \multirow{1}{*}{Table height} & - & - & $[-0.025, 0.025]$ & $[-0.025, 0.025]$ \\
    \bottomrule
\end{tabular}
\label{tab:sim_variation}
\end{table}

\textbf{Pick-Place}: The robot must pick up a round toy tomato and place it onto a metal plate. Success is defined as the tomato is within $5cm$ to the center of the plate. For this task, we collect training data by replaying real-world trajectories of the real \emph{Pick Place} task. We use two $192\times192$ RGB cameras: one mounted on the wrist and one positioned off‑table, pointing at the table center. The initial dataset contains 150 demonstrations, with 30 demos per factor.

\textbf{Peg Insertion}: The robot must pick up a rectangular peg and insert it into a hole in a box, requiring high precision. Success is defined as half of the peg is inserted into the hole. Two $256\times256$ RGB cameras are used: a wrist camera and a third‑person‑view camera positioned off‑table, pointing at the table center. We adapt this task from the ManiSkill3 codebase \citep{taomaniskill3} and use a scripted policy to collect data. For the visual task, the initial dataset includes 150 demos (30 per factor); for the spatial task, it includes 240 demos (30 per factor).

\textbf{Pull Cube Tool}: The robot must first pick up an L‑shaped tool and then use it to pull a cube closer, beyond its unaided reach. Success is defined as pulling the cube to within $45,\text{cm}$ of the robot base. One $192\times192$ RGB camera is placed off‑table, pointing at the table center. We adapt this task from the ManiSkill3 codebase \citep{taomaniskill3} and use a scripted policy to collect data.  For the visual task, the initial dataset contains 150 demos (30 per factor); for the spatial task, it contains 240 demos (30 per factor).

\subsection{Policy Implementation Details}
\label{app:policy_details}

All policies are trained with Diffusion Policy \cite{chi2023diffusionpolicy}. We use ResNet-18~\cite{he2015deepresiduallearningimage} as our vision encoder. Each policy undergoes 50000 gradient updates with a fixed batch size of 64, yielding identical computational cost across datasets of different sizes. RGB observations are augmented with standard color‑jitter during training. A complete list of hyper‑parameters is provided in \cref{table:app_hyperparam_sim_dp}.

The robot state is an 8‑dimensional vector comprising the seven joint positions and a single gripper state. Actions are specified as 8‑dimensional absolute joint‑position commands sent to a absolute position controller.

\begin{table*}[h]
\centering
\small
\renewcommand{\arraystretch}{1.05}
\caption{Hyper-parameters of simulation diffusion policy.}
\begin{tabular}{
  c
  c
  c
  c
  c
  c
  c
}
\toprule
 Model Dimension & Dim Mults & Time Embedding Dimension & History Steps & Horizon & Action Steps \\
\midrule
\addlinespace
128 & [1,2,4]  & 128  & 1 & 16 & 8 \\
\bottomrule
\end{tabular}
\label{table:app_hyperparam_sim_dp}
\end{table*}

\subsection{Evaluation Details}
Each policy is evaluated on ten discrete settings per factor, different from the training settings. For every setting we execute 60 trials with distinct initial states, resulting in $N\times10\times60$ rollouts---3000 trials in the visual‑factor regime and 4800 trials in the full‑factor regime. Reported success rates are the mean over all rollouts.

\section{Real Robot Experiment}
\label{app:real_exp_setup}
\subsection{Hardware Setup}

We use a Franka Panda robot for our real robot experiment. We use Logitech C920 webcam as our third person camera, and RealSense D405 for the wrist camera. Both cameras use resolution $192\times192$. We use a Meta Quest 2 VR headset for teleoperation to perform data collection.

\begin{figure*}[htbp]
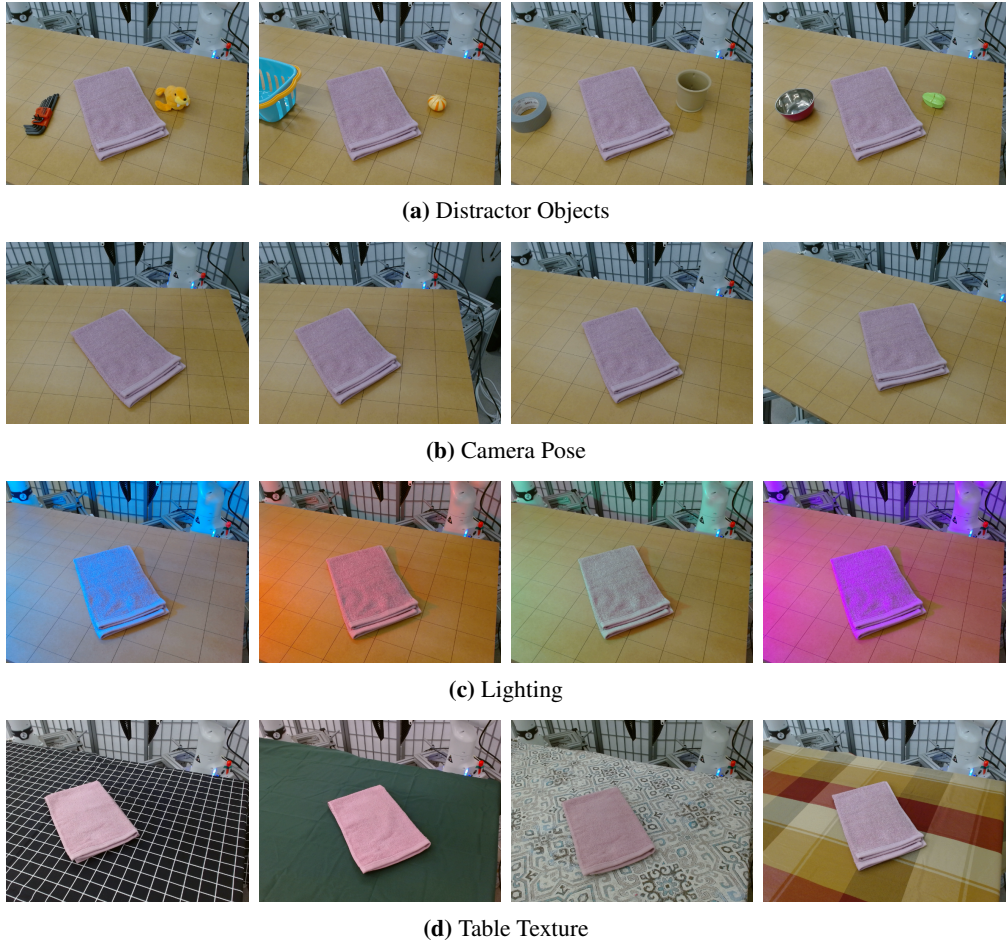

\centering
\newcommand{\simvar}[2]{\includegraphics[width=0.23\textwidth]{figures/real_variation/#1_#2.png}}

\begin{subfigure}[t]{\textwidth}
    \centering
    \simvar{distractor}{0}\hspace{4pt}%
    \simvar{distractor}{1}\hspace{4pt}%
    \simvar{distractor}{2}\hspace{4pt}%
    \simvar{distractor}{3}
    \caption{Distractor Objects}
    \vspace{5pt}
\end{subfigure}

\begin{subfigure}[t]{\textwidth}
    \centering
    \simvar{camera_pose}{0}\hspace{4pt}%
    \simvar{camera_pose}{1}\hspace{4pt}%
    \simvar{camera_pose}{2}\hspace{4pt}%
    \simvar{camera_pose}{3}
    \caption{Camera Pose}
    \vspace{5pt}
\end{subfigure}

\begin{subfigure}[t]{\textwidth}
    \centering
    \simvar{lighting}{0}\hspace{4pt}%
    \simvar{lighting}{1}\hspace{4pt}%
    \simvar{lighting}{2}\hspace{4pt}%
    \simvar{lighting}{3}
    \caption{Lighting}
    \vspace{5pt}
\end{subfigure}

\begin{subfigure}[t]{\textwidth}
    \centering
    \simvar{table_texture}{0}\hspace{4pt}%
    \simvar{table_texture}{1}\hspace{4pt}%
    \simvar{table_texture}{2}\hspace{4pt}%
    \simvar{table_texture}{3}
    \caption{Table Texture}
\end{subfigure}

\caption{\small{Visualization of real environment factor variations.}}
\label{fig:real_variation_factors}
\end{figure*}

\subsection{Task and Factor Description}

For training, we sample four pre-specified camera poses for the third-person camera, as visualized in \cref{fig:real_variation_factors}. We use four textured and colored cloths to set up table texture variations. We use four sets of distractors for the distractor factor variation, where each set of distractor contains two objects, e.g., bread, eggplant, grape, carrot, etc. For spatial factor experiments, robot initial joint position is drawn from [-0.015,+0.015] around its nominal joint positions. Table height is omitted because it is difficult to change in our real world experiment setting. We increase the range of object pose by $25\%$ more for object pose variation. We visualize the visual factor variations in \cref{fig:real_variation_factors}. 

\textbf{Pick place}: the robot needs to pick up a round tomato and place it into a metal plate. The tomato position and the plate position and randomly set in a $40cm \times 40cm$ grid. The rotation of the plate is randomly set across training demonstrations and evaluations. We consider an initial dataset size of 120 demos, where we have 30 demos for each factor.

\textbf{Fold Towel}: the robot needs to grasp the end of a rectangular towel and fold it in half across the line bisecting the longer side. We collect training data with the towel position randomly set in a $5cm \times 5cm$ grid and rotation between $30^\circ \text{ to } 60^\circ$ counterclockwise relative to the vertical axis. For \emph{Fold Towel - Visual}, we consider an initial dataset size of 120 demos, where we have 30 demos for each factor. For \emph{Fold Towel - Spatial}, we consider an initial dataset size of 180 demos, where we have 30 demos for each factor.

\textbf{Mouse in Drawer}: the robot needs to open a drawer, pick up a mouse, place it in the opened drawer, and close the drawer. We collect training data with drawer and mouse positions each randomly set within $\sim 10\,\text{cm}$ and rotations within $\pm 10^\circ$ of a fixed initial setup. We consider an initial dataset size of 180 demos, where we have 30 demos for each factor.

\subsection{Policy Implementation Details}

For \emph{Pick Place} task, we use diffusion policy \cite{chi2023diffusionpolicy} to train all the policies. We follow the same color jitter augmentation protocol and hyper-parameters in \cref{table:app_hyperparam_sim_dp}.

For \emph{Fold Towel} and \emph{Mouse in Drawer} task, we fine-tune $\pi_0$ on our collected dataset. Specifically, we fine-tune from $\pi_0-base$ model. We freeze the ViT and the language model, and only train the action expert. We train all policies for 10,000 gradient steps for the same batch size 32, resulting in an equal training cost regardless of dataset size.

We use absolute joint position control for all the tasks. The input to the policy is camera images and a 8-dimensional state vector, consisting of robot current joint angles and gripper state. The output is a 8-dimensional vector, consisting of robot target joint angles and target gripper state. The control frequency is 15 Hz.

\subsection{Evaluation Details}

We evaluate each policy in difficult out-of-distribution cases where we randomly draw values for each $f_i$ different from the training environment. 

For the \emph{Pick Place} task, we evaluate each policy on 10 factor value combinations, 2 trials per combination, for 20 trails in total. We assign $0/1$ success. 

For \emph{Fold Towel} task, we evaluate each policy on 4 factor value combinations, 3 trials per value, for 12 trails in total. We assign partial credit, where $0$ stands for complete failure, $0.25$ stands for underfold/overfold by more than 5 centimeters or more than $20^\circ$, $0.5$ stands for underfold/overfold by less than 5 centimeters and less than $20^\circ$ but more than $3cm$ or $5^\circ$, and $1$ for complete success. 

For \emph{Mouse in Drawer} task, we evaluate each policy on 6 factor value combinations, 3 trials per value, for 18 trails in total. We assign $0$ for failing to open the drawer or pick up the mouse, $0.25$ for successfully picking up the mouse and failing to put in the drawer, $0.5$ for successfully putting the mouse into the drawer but failing to close the drawer, $1$ for complete success.

The rollout is terminated early if the robot collides with the table or enters any other hazardous state, and the trial is marked as a failure. Each rollout is capped at 600 environment steps; any trial that exceeds this limit is recorded as a failure.

\subsection{Baseline Details}
\label{app:baseline}

\paragraph{Re-Mix.} We train a discrete reference model with domain weights proportional to size and select the best reference model by lowest validation loss. Next, we learn the domain weights by applying robust optimization that minimizes worst case excess loss between the learned and reference policy. We take the average value of the domain weights across robust optimization training and use it for downstream policy training. 

\begin{table}[htbp]
  \centering
  \caption{Hyperparameters: Remix}\label{tab:hyperparams-remix}
  \vspace{0.1in}

  \scriptsize                              
  \setlength{\tabcolsep}{3pt}              
  \renewcommand{\arraystretch}{1.05}       

  \resizebox{\linewidth}{!}{%
  \begin{tabular}{@{}lll@{}}
    \toprule
    \textbf{Group} & \textbf{Hyperparameter} & \textbf{Value} \\ \midrule
    \textbf{Dataloader}         & batch size            & 32 \\ \midrule
    \textbf{Action Head (Reference)}
                                & head type             & \texttt{DDPMActionHead} \\
                                & model class           & \texttt{ConditionalUnet1D} \\
                                & down features         & (256, 512, 1024) \\
                                & mid layers            & 2 \\
                                & time features         & 128 \\
                                & kernel size           & 5 \\
                                & clip sample           & 1.0 \\
                                & diffusion timesteps   & 100 \\
                                & variance type         & fixed small \\ \midrule
    \textbf{Action Head (Remix)}
                                & head type             & \texttt{DiscreteActionHead} \\
                                & model class           & \texttt{MLP} \\
                                & hidden dims           & (512, 512, 512) \\
                                & dropout rate          & 0.4 \\
                                & activate final layer  & True \\
                                & layer normalization   & True \\
                                & number of action bins & 48 \\
                                & bin type              & gaussian \\ \midrule
    \textbf{LR Schedule} (\texttt{optax.warmup\_cosine\_decay\_schedule})
                                & initial value         & $1\times10^{-6}$ \\
                                & peak value            & $1\times10^{-4}$ \\
                                & warm-up steps         & 1 000 \\
                                & decay steps           & 500 000 \\
                                & end value             & $1\times10^{-6}$ \\ \midrule
    \textbf{Training / DoReMi}  & domain-weight step size & 0.2 \\
                                & smoothing             & $5\times10^{-2}$ \\ \bottomrule
  \end{tabular}}%
\end{table}



\end{document}